\documentclass[twoside]{article}

\usepackage[accepted]{aistats2023}
\usepackage[american]{babel}
\usepackage[authoryear,round]{natbib}
\usepackage{hyperref}
\usepackage{mathtools} % amsmath with fixes and additions
\usepackage{amssymb}
\usepackage{booktabs} % commands to create good-looking tables

\newcommand{\R}{\mathbb{R}}

\newcommand{\abs}[1]{\left\lvert#1\right\rvert}
\newcommand{\norm}[1]{\left\lVert#1\right\rVert}

\begin{document}

% If your paper is accepted and the title of your paper is very long,
% the style will print as headings an error message. Use the following
% command to supply a shorter title of your paper so that it can be
% used as headings.
%
\runningtitle{Encoding Domain Knowledge in Multi-view Latent Variable Models}

% If your paper is accepted and the number of authors is large, the
% style will print as headings an error message. Use the following
% command to supply a shorter version of the authors names so that
% they can be used as headings (for example, use only the surnames)
%
%\runningauthor{Surname 1, Surname 2, Surname 3, ...., Surname n}

\twocolumn[

\aistatstitle{Encoding Domain Knowledge in Multi-view Latent Variable Models:\\ A Bayesian Approach with Structured Sparsity}

\aistatsauthor{ Arber Qoku \And Florian Buettner }

\aistatsaddress{
German Cancer Research Center (DKFZ) \\
German Cancer Consortium (DKTK) \\
Goethe University Frankfurt, Germany \\
\texttt{arber.qoku@dkfz.de} \\ 
\And  
German Cancer Research Center (DKFZ) \\
German Cancer Consortium (DKTK) \\
Frankfurt Cancer Institute, Germany \\
Goethe University Frankfurt, Germany \\
\texttt{florian.buettner@dkfz.de} \\ 
} ]

\begin{abstract}
Many real-world systems are described not only by data from a single source but via multiple data views. In genomic medicine, for instance, patients can be characterized by data from different molecular layers. Latent variable models with structured sparsity are a commonly used tool for disentangling variation within and across data views. However, their interpretability is cumbersome since it requires a direct inspection and interpretation of each factor from domain experts. Here, we propose MuVI, a novel multi-view latent variable model based on a modified horseshoe prior for modeling structured sparsity. This facilitates the incorporation of limited and noisy domain knowledge, thereby allowing for an analysis of multi-view data in an inherently explainable manner. We demonstrate that our model (i) outperforms state-of-the-art approaches for modeling structured sparsity in terms of the reconstruction error and the precision/recall, (ii) robustly integrates noisy domain expertise in the form of feature sets, (iii) promotes the identifiability of factors and (iv) infers interpretable and biologically meaningful axes of variation in a real-world multi-view dataset of cancer patients.
% Code available at \href{https://github.com/MLO-lab/MuVI}{github.com/mlo-lab/muvi}.
\end{abstract}

\section{INTRODUCTION}\label{sec:intro}
\begin{figure}[t]
  \centering
  \vspace{.3in}
  \includegraphics[width=0.85\linewidth]{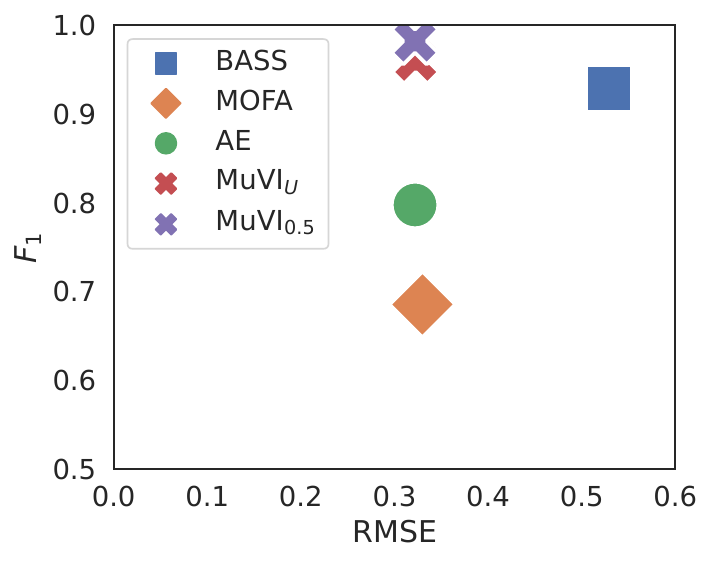}
  \vspace{.2in}
  \caption{Performance of our model against competitive models. MuVI achieves a higher $F_1$ score for recovering true active features while maintaining a low reconstruction error, even when using an uninformative prior (MuVI$_U$).}
  \label{fig:teaser_new}
\end{figure}

In many real-world applications, complex systems are characterized via multiple data views. That is, observations are represented by multiple groups of distinct features. These groups of features often describe different and complementary input sources that are required for a comprehensive characterization of a sample. For example, in genomic medicine, a single patient can be described by quantifying different molecular layers such as the proteome, the microbiome and the transcriptome.\\
Latent variable models are powerful statistical tools that uncover the axes of variation between samples and data views, by inferring unobserved hidden states from the observable high-dimensional data. To disentangle the sources of heterogeneity driving intra-view and inter-view variation in a meaningful manner, it is key that latent variable models are expressive and interpretable. 
Modeling approaches with high expressive power are based on autoencoders~\citep{ainsworth2018oi} or Gaussian Process Latent Variable models~\citep{damianou2012manifold}. However, the non-linearities introduced to model complex dependencies hinder interpretability to the extent that linear approaches remain the dominant tool for analyzing multi-view data~\citep{argelaguet2018multi}. Recently proposed AE-based approaches attempt to balance expressive power and interpretability by combining non-linear encoders with linear decoders~\citep{svensson2020interpretable, lotfollahi2023biologically}.\\
Commonly used approaches that yield an interpretable decomposition of multi-view data are factor analysis models with structured sparsity on the view, factor and feature level. This modeling task is challenging since the priors inducing the structured sparsity should (i) be amenable to efficient inference strategies that scale to large datasets, (ii) identify true active features and factors across views with a high precision and recall,
and (iii) faithfully model the data with a low reconstruction error. Current state-of-the-art methods typically achieve structured sparsity by incorporating a spike-and-slab prior~\citep{argelaguet2018multi} or a three-parameter beta (TPB) prior~\citep{zhao2016bayesian}. 
While these models result in sparse solutions, a careful inspection of each latent factor is necessary to provide a meaningful interpretation that is relevant to the domain of study. 
% These models result in sparse factor loadings, allowing for a direct inspection and interpretation of each factor from a domain expert.\\
This manual annotation is cumbersome, time consuming and requires highly-specialized expertise. In addition, the inferred latent representations cannot be directly compared across several training instances as these approaches are unidentifiable unless additional constraints are put on their latent components~\citep{anderson1956statistical}. In brief, non-identifiability in factor analysis states that multiple solutions can produce identical likelihoods, e.g.\ by applying orthogonal transformations on the latent components, or by permuting the factor indices.\\
% For instance, multiplying the factor loadings with an orthogonal matrix and multiplying the factor scores with the transposed orthogonal matrix leads to identical solutions. Similarly, scaling the factor loadings by a constant and downscaling the factor scores by the inverse of that constant. Similarly, applying the same permutation to the factor indices of both the factor scores and the factor loadings leads to identical solutions.\\
To automate a consistent annotation process, we propose to leverage the partial knowledge on the structure of the latent space that is readily available %in a structured manner
in many domains. For example, in genomic medicine, disease states are often characterized by sets of features (or pathways), comprising genes that are known to act in a coordinated manner. These data, collected over many years of scientific progress, are curated in dedicated pathway databases.
Integrating this domain knowledge in a principled manner as sparsity priors in multi-view models is challenging. Domain knowledge is noisy (e.g. pathways contain many false positive and false negative annotations) and domain knowledge is incomplete and often only available for a subset of views (e.g. in genomic medicine pathway information is only readily available for the transcriptome and the genome).\\
In this work we propose a novel \textbf{mu}lti-\textbf{v}iew latent variable model with domain-\textbf{i}nformed structured sparsity (MuVI) for incorporating domain knowledge via structured sparse priors, and analyzing multi-view data in an inherently explainable manner. We first introduce a modified horseshoe prior for inducing structured sparsity that (i) maintains a low reconstruction error while identifying true active features more reliably than state-of-the-art sparse priors, and (ii) thereby facilitates the integration of domain knowledge from noisy feature sets. Our domain-informed priors render the inferred latent variables directly interpretable without any further interaction with domain expert, by tagging each latent variable with its corresponding feature set (or pathway in genomic medicine). Briefly, the contributions of our work are as follows:
\begin{itemize}
    \item We demonstrate in a comprehensive evaluation that our model outperforms state-of-the-art approaches in terms of the reconstruction error and precision, recall and $F_1$ score.
    \item We show that our model utilizes prior information efficiently by recovering correct signals from noisy feature sets and is robust against poorly specified priors.
    \item We demonstrate the practical utility of our model on a large multi-view dataset of cancer patients by inferring interpretable and biologically meaningful axes of variation.
\end{itemize}

\section{RELATED WORK}\label{sec:related}

Factor analysis (FA) is a fundamental approach for estimating and understanding the correlation structure among observed variables~\citep{thurstone1931multiple}, which has inspired the development of numerous latent variable models. Due to its simplicity, however, standard FA is unable to model observations from multiple sources. Extensions, such as the canonical correlation analysis (CCA)~\citep{hotelling1992relations, klami2013bayesian}, or the group factor analysis (GFA)~\citep{klami2014group} model paired observations simultaneously by learning linear dependencies underlying two or more data sources. A central assumption of the GFA is that the multi-view observations are a manifestation of a lower dimensional common latent space corrupted by Gaussian noise. The main components that describe this relationship are the factor-to-feature linear mappings or the so-called factor loadings. Factor loadings encode the structure of each factor, and play an important role in the interpretation of the model. Hence, factor loadings are suitable for introducing statistical assumptions on their underlying structure such as sparsity. For instance, GFA quantifies the association between view and factor by extending the automatic relevance determination~\citep{mackay1994bayesian}. 

\subsection{Sparsity Inducing Priors}
In non-Bayesian approaches, sparsity is commonly handled by introducing additional terms to the optimization objective posed on the model parameters. A canonical example is the L1 penalty or lasso~\citep{tibshirani1996regression}.
On the other hand, Bayesian approaches achieve sparse solutions via sparsity inducing priors. For instance, the double exponential, or Laplace prior is the Bayesian counterpart of the lasso~\citep{park2008bayesian}. Another popular choice is the discrete spike-and-slab prior~\citep{mitchell1988bayesian}, a mixture of a Dirac delta distribution centered at zero for pruning irrelevant signals, and a normal distribution for modeling larger signals. Recently, the spike-and-slab lasso (SSL)~\citep{rovckova2018spike} has emerged as a combination of two Laplace distributions that are parameterized to emulate both the spike and the slab component. 
Other flavors of shrinkage priors~\citep{polson2010shrink} include the horseshoe prior~\citep{carvalho2009handling}, which offers a continuous relaxation of the spike-and-slab approach, thereby preserving sparse properties while providing computational benefits such as being differentiable.

\subsection{Structured Sparsity in Multi-view LVMs}
Several Bayesian approaches have successfully utilized the synergy between latent variable models and sparsity inducing priors~\citep{bernardo2003bayesian, engelhardt2010analysis, lan2014sparse}.
In the multi-view setting, \citep{zhao2016bayesian} propose a hierarchical Bayesian GFA with structured sparsity, facilitated by a cascading three parameter beta prior~\citep{armagan2011generalized}. The consequences of this addition are twofold, the column-wise sparsity supports the inference of associations between views, while element-wise sparsity encourages feature selection within individual factors. Multi-omics factor analysis (MOFA)~\citep{argelaguet2018multi} assumes a similar structured sparsity in two levels, which is achieved by combining an automatic relevance determination (ARD) prior~\citep{mackay1994bayesian} with a spike-and-slab prior. Beyond linear models, there is an array of non-linear modeling approaches that generalize the ideas of linear LVMs and sparse priors to more complex data settings. These include approaches based on variational autoencoder~\citep{ainsworth2018oi} or Gaussian Process Latent Variable Models~\citep{damianou2012manifold}. Such approaches, however, are difficult to interpret in practice and have therefore not received widespread application. For example, in the latter GP-based approach, loading matrices are marginalized out, and cannot be used to identify any physical or biological processes captured by individual factors. Recent hybrid approaches~\citep{svensson2020interpretable, lotfollahi2023biologically} combine non-linear encoders with linear decoders to maintain the expressive power of autoencoders while preserving the interpretability of linear models. While these approaches were proposed for single data views, extension to the multi-view setting is straightforward.

\subsection{Integration of Domain Knowledge in LVMs}

Typically, before training sparse latent variable models, the prior is set to be relatively uninformative, allowing only the observations to mold the posterior. An alternative approach is introducing an informed structured sparsity into the model based on external expertise in the domain of interest. The factorial single-cell latent variable model~\citep{buettner2017f} attempts to bridge this gap by explicitly integrating, and jointly modeling a collection of domain relevant feature sets such as gene set annotations. The model extends the spike-and-slab to be informed by the presence or absence of a feature, and infers pre-labeled latent axes of variation. However, this method can only handle a single data view and the proposed inference scheme does not scale to large sample sizes. An orthogonal approach for identifying latent variables corresponding to pathways is the single sample gene set enrichment analysis (ssGSEA)~\citep{subramanian2005gene,barbie2009systematic}, 
which computes a sample-wise enrichment of a pathway. 
% where an absolute enrichment of a pathway in each sample is computed. 
However, this approach does not account for noisy annotations, treats pathways as independent and can only be applied to a single data view.
A multi-view approach with structured sparsity that admits prior information in terms of feature sets, and handles partial and potentially noisy priors has not yet been explored.

\section{DOMAIN-INFORMED MULTI-VIEW MODELING}\label{sec:method}
\subsection{Background and Notation}\label{subsec:background}
% maybe add a few more subsections...first GFA
Let $\mathbf{y}_i\in\R^D$ denote a $D$-dimensional observation for $i \in \left\{1,\dots,N\right\}$, and $G_m \subseteq \{1,\dots,D\}$ describe a mutually disjoint grouping of the features into $M$ data views, where $G_p \cap G_q = \emptyset$ for $p \neq q \in \{1,\dots,M\}$. For simplicity, we assume a permutation of the features $D$ such that the first $D_1$ features belong to $G_1$, the second $D_2$ features to $G_2$ and so on. As a result, we may rewrite the collection of the observations as a matrix $\mathbf{Y} \in \R^{N\times D}$ comprising $M$ coupled views $\left[\mathbf{Y}^{(1)}, \mathbf{Y}^{(2)},\dots,\mathbf{Y}^{(M)}\right]$, where $\mathbf{Y}^{(m)} \in \R^{N\times D_m}$. The main goal is to then represent each observation $\mathbf{y}_i$ in terms of a low-dimensional set of latent factors $\mathbf{x}_{i}\in\R^K, K \ll D$. The relationship between the observations and the latent factors is described by a set of view-specific factor loadings $\mathbf{W}^{(m)}\in\R^{D_m \times K}$. Then, the recipe for the group factor analysis framework posits the following generative process to the observed data,
\begin{equation}\label{eq:gfa}
\mathbf{y}_i^{(m)} \sim \mathcal{N}\left(\mathbf{W}^{(m)}\mathbf{x}_i, \mathbf{\Psi}^{(m)}\right)
\end{equation}
where each latent variable typically follows an isotropic standard normal distribution 
\begin{equation}
    \mathbf{x}_i \sim \mathcal{N}\left(\mathbf{0}, \mathbf{I}\right).
\end{equation}
The residuals are denoted by $\mathbf{\Psi}^{(m)} = diag(\mathbf{\sigma}^{2(m)})$, a diagonal matrix storing the marginal variances $\sigma_j^{2(m)}$ of each variable $j$ in view $m$. Due to conjugacy properties, setting 
\begin{equation}
    \sigma_j^{2(m)} \sim \Gamma^{-1}(\alpha_{\sigma}, \beta_{\sigma})
\end{equation}
is a common choice, where $\Gamma^{-1}(\alpha, \beta)$ describes the inverse-Gamma with shape and scale parameters $\alpha$ and $\beta$.
Finally, an important component of GFA is the collection of the factor loadings $\mathbf{W}^{(m)}$ which applies a linear projection of the latent variable $\mathbf{x}_i$ to $\mathbf{y}_i^{(m)}$.
In order to facilitate the interpretability of the factor-to-feature mapping, \citep{klami2014group} propose a structured sparsity for the columns of $\mathbf{W}^{(m)}$, such that each factor falls into two distinct categories. A non-zero factor loading vector $\mathbf{w}_{k}^{(m)}$ indicates an active factor $k$, while $\mathbf{w}_{k}^{(m)} = \mathbf{0}$ indicates an inactive factor $k$ in view $m$. Hence, a factor is either shared across an arbitrary subset of views, or private to a specific view.

\subsection{MuVI}\label{subsec:MuVI}
We follow a similar approach, inspired by the success of the horseshoe prior~\citep{carvalho2009handling, carvalho2010horseshoe}, and introduce a \emph{view-factor-local} shrinkage prior on the factor loadings to enable both column-wise and element-wise shrinkage:
\begin{equation}\label{eq:hs}
w_{j, k}^{(m)} \sim \mathcal{N}\left(0, \left(\tau^{(m)} \delta_{k}^{(m)} \lambda_{j, k}^{(m)}\right)^2\right),
\end{equation}
where each scale in the hierarchy follows a positive Cauchy distribution, 
\begin{align}
    \tau^{(m)}&\sim\mathcal{C}^{+}(0, 1)\\
    \delta_{k}^{(m)}&\sim\mathcal{C}^{+}(0, 1)\\
    \lambda_{j, k}^{(m)}&\sim\mathcal{C}^{+}(0, 1).
\end{align}
Each level in the hierarchy contributes to the overall structured sparsity of the factor loadings. In particular, $\delta_{k}^{(m)}$ serves as an automatic relevance determination (ARD) mechanism~\citep{mackay1994bayesian} for factor $k$ in view $m$, effectively decoupling this factor from the rest of the views. At the same time, $\lambda_{j, k}^{(m)}$ acts as a regulator on each individual loading, encouraging the model to seek simpler solutions that describe each factor in terms of fewer features. However, due to the heavy tails of the Cauchy distribution, weakly identified loadings under the horseshoe prior can easily escape the regularization. To counteract this behavior, the regularized horseshoe guarantees a non-zero penalty even for large weights~\citep{piironen2017sparsity}. Concurrently, the regularized parameterization helps integrate prior information about the structure of the latent factors into the model. We update Equation~\ref{eq:hs} as follows. To simplify notation, we sometimes drop the view-specific superscript $m$, which can also be implicitly encoded in $j$. Let $\gamma_{j, k} = \tau \delta_{k} \lambda_{j, k}$, then
\begin{equation}\label{eq:reg_hs}
w_{j, k} \sim \mathcal{N}\left(0, \frac{\left(c_{j, k}\gamma_{j, k}\right)^2}{c_{j, k}^2 + \gamma_{j, k}^2}\right)
\end{equation}
Note that in Equation~\ref{eq:reg_hs} the $c_{j, k}^2$ parameter relates to a specific factor loading rather than being a global parameter as originally defined in the regularized horseshoe. For a relatively large $c_{j, k}^2 \gg \gamma_{j, k}^2$, $w_{j, k}$ is scaled by a factor of nearly $\gamma_{j, k}$, rendering the effect of the additional penalty insignificant, and reinstating the original horseshoe prior. On the other hand, when $c_{j, k}^2 \ll \gamma_{j, k}^2$, $w_{j, k}$ is scaled by a factor of nearly $c_{j, k}$, thus setting an upper bound on corresponding factor loadings. Equivalently, the regularized horseshoe can be seen as a continuous alternative to the discrete spike-and-slab prior~\citep{mitchell1988bayesian} with a finite slab width. \cite{piironen2017sparsity} suggest assigning an inverse-Gamma distribution as a weakly informative prior to
\begin{equation}
    c_{j, k}^2 \sim \Gamma^{-1}(\alpha_{c}, \beta_{c}).
\end{equation}
For all our experiments we opt for $\alpha_{c} = \beta_{c} = 0.5$, as this encourages sparsity while still allowing strong signals to escape the regularization due to the heavy right tail.

\subsection{Integrating Prior Knowledge from Noisy Feature Sets}\label{subsec:prior}
Next, we attempt to integrate prior knowledge in terms of noisy feature sets into our model.
\paragraph{Feature sets}
A feature set $\mathbf{I}^D$ is a collection of binary variables $I_j\in\left\{0, 1\right\}$, $j \in \left\{1,\dots,D\right\}$, where $I_j = 1$ indicates the presence of feature $j$, and $I_j = 0$ its absence. Assume we have substantial knowledge about the underlying structure of the factor loadings in terms of feature sets. That is, for every latent dimension $k$ and a set of features $D$ we are given a corresponding feature set $\mathbf{I}_k^D$, such that $\abs{w_{j, k}} > 0$ if $I_{j, k} = 1$, and $w_{j, k} = 0$ otherwise. %Then, the solution to a sparse group factor analysis is trivial.
Each latent factor can be seen as a factor-to-feature mapping of the present features determined by the feature set. In terms of Equation~\ref{eq:reg_hs}, this can also be achieved by setting the corresponding slab width $c_{j, k} \approx 0$, i.e.\ applying an infinitely large weight decay penalty to $w_{j, k}$.
\paragraph{Noisy feature sets}
In practice, we rarely have access to such pristine ground truth. However, in some cases, we may exploit existing domain knowledge to develop a prior belief about a plausible structure of the factor loadings by accommodating noisy feature sets into our modeling approach. Let $\tilde{\mathbf{I}}^D$ be a noisy version of a feature set $\mathbf{I}^D$, where a subset of the binary variables $I_{Q}, Q \subset \left\{1,\dots,D\right\}$ has been flipped to generate $\tilde{\mathbf{I}}^D$, inserting a non-zero fraction of false positives and false negatives. To integrate noisy feature sets into MuVI, we relax the hard regularization penalty induced by a pre-defined $c_{j, k}$, allowing the adaptation of incorrect signals, given sufficient evidence from the data. In our approach, we opt for an auxiliary hyperparameter $0 < \alpha_{j, k} \leq 1$ which scales $c_{j, k}$. A value of $\alpha_{j, k} = 1.0$ poses no prior penalty to $w_{j, k}$, while $\alpha_{j, k} < 1.0$ leads to a smaller slab width a priori. Consequently, a smaller $\alpha_{j, k}$ translates to a stronger prior belief. Experiments show that values around $0.01 \leq \alpha_{j, k} \leq 0.05$ for absent features in a prior collection of feature sets consistently provide the best results across different datasets and training scenarios.

\newpage
\paragraph{Pure variables}
Present features that depend on only one latent factor are also referred to as pure variables~\citep{bing2020adaptive}. Work on the identifiability of factor analysis~\citep{bing2020adaptive,anderson1956statistical} proves that the existence of at least two pure variables is a sufficient condition for yielding identifiable solutions. That is, for every factor $k\in\{1,\dots, K\}$, there exist at least two features $j, j'\in\{1,\dots, D\}$ such that $I_{j, k} = 1, I_{j', k} = 1$ for all $j\neq j'$. Under the assumption of known pure variables, MuVI results in provably identifiable factors. However, since we are only given a noisy feature set, we cannot know whether pure variables really are present for all factors. We therefore investigate empirically to what extent our model preserves the identifiability properties when increasing the amount of noise in the prior annotations.

\subsection{Inference}\label{subsec:inference}
The joint model can be written as
\begin{align}
    p\left(\mathbf{Y}, \mathbf{\Theta}\right)
    &= p\left(\mathbf{Y}, \mathbf{X}, \mathbf{W}, \mathbf{\Lambda}, \mathbf{\Delta}, \mathbf{\tau}, \mathbf{C}, \mathbf{\Psi}\right)\nonumber\\
    &= p\left(\mathbf{Y} \mid \mathbf{X}, \mathbf{W}, \mathbf{\Psi}\right)p\left(\mathbf{\Psi}\right)\nonumber\\
    &\times p\left(\mathbf{X}\right)p\left(\mathbf{W} \mid \mathbf{\Lambda}, \mathbf{\Delta}, \mathbf{\tau}, \mathbf{C}\right)\nonumber\\
    &\times p\left(\mathbf{\Lambda}\right)p\left(\mathbf{\Delta}\right)p\left(\mathbf{\tau}\right)p\left(\mathbf{C}\right),
\end{align}
where $\mathbf{\Theta} =\mathbf{X}, \mathbf{W}, \mathbf{\Lambda}, \mathbf{\Delta}, \mathbf{\tau}, \mathbf{C}, \mathbf{\Psi}$, and
\begin{align*}
    \mathbf{Y} &\!=\! \left\{y_{i, j}^{(m)}\right\},
    \mathbf{\Psi} \!=\! \left\{\sigma_{j}^{(m)}\right\},
    \mathbf{X} \!=\! \left\{x_{i, k}\right\},
    \mathbf{W} \!=\! \left\{w_{j, k}^{(m)}\right\},\\
    \mathbf{\Lambda} &\!=\! \left\{\lambda_{j, k}^{(m)}\right\},
    \mathbf{\Delta} \!=\! \left\{\delta_{k}^{(m)}\right\},
    \mathbf{\tau} \!=\! \left\{\tau^{(m)}\right\},
    \mathbf{C} \!=\! \left\{c_{j, k}^{(m)}\right\}.
\end{align*}
Inference is performed by introducing a fully factorized family of parameterized distributions $q_{\phi}\left(\mathbf{\Theta}\right) = \prod_{\theta \in \mathbf{\Theta}} q_{\phi}\left(\theta\right) $ to approximate the intractable posterior $p\left(\mathbf{\Theta}\mid\mathbf{Y}\right)$. We maximize the evidence lower bound with respect to the variational parameters $\phi$, which, in return reduces the gap between the true and the approximate posterior in terms of the KL divergence. The family of normal distributions is a natural choice for approximating $\mathbf{X}$ and $\mathbf{W}$, 
% that is, $q_{\phi_{x}}(x_{i, k}) = \mathcal{N}(\mu_x, \sigma_x^2)$ and $q_{\phi_{w}}(w_{i, k}^{(m)}) = \mathcal{N}(\mu_w, \sigma_w^2)$. 
whereas, for the rest of the parameters we assert the log-Normal distribution to ensure positive samples~\citep{ghosh2018structured}. 
The resulting optimization objective 
\begin{equation}
    \mathcal{L}(\phi) = \mathbb{E}_{q_{\phi}}\left[\log p(\mathbf{Y}, \mathbf{\Theta}) - \log q_{\phi}\left(\mathbf{\Theta}\right)\right]
\end{equation}
is amenable to automated stochastic variational inference~\citep{ranganath2014black, hoffman2013stochastic}, which involves sampling from the variational distribution, and taking unbiased but noisy Monte Carlo estimates of the gradient $\nabla_{\phi}\mathcal{L}$. In addition, the reparameterization trick~\citep{kingma2013auto} further stabilizes the optimization procedure, in which the random variables are expressed as a combination of deterministic variables and external random noise, thereby greatly reducing the variance of the MC estimates. 
We provide an implementation of MuVI using Pyro~\citep{bingham2019pyro} on GitHub\footnote{\href{https://github.com/MLO-lab/MuVI}{https://github.com/MLO-lab/MuVI}}.

\section{EXPERIMENTS}\label{sec:exp}
We first evaluate different aspects of MuVI empirically on a wide range of simulated settings. Ideally, the model performs well in the following general tasks:
\begin{itemize}
    \item Learning a meaningful latent representation with low reconstruction loss - both in the presence and absence of prior information.
    \item Utilizing prior information efficiently to recover correct signals from noisy feature sets, and being robust against poorly specified or entirely incorrect priors.
    % \item Being robust against poorly specified or entirely incorrect prior information.
    \item Promoting an implicit flow of the prior information from the informed to the uninformed views via the shared factors.
\end{itemize}

We compare our model against four baselines that admit observations across multiple views, and learn a sparse representation of the common latent space: the Bayesian group factor analysis with structured sparsity (BASS)~\citep{zhao2016bayesian}, multi-omics factor analysis (MOFA)~\citep{argelaguet2018multi} and a multi-view autoencoder (AE), as a naive multi-view extension of the interpretable autoencoder~\citep{svensson2020interpretable}. We also include the group factor analysis (GFA)~\citep{klami2014group} for the synthetic evaluation, while keeping in mind that GFA does not infer sparse solutions regarding the factor loadings. Finally, we demonstrate the practical utility of MuVI on a real-world dataset from The Cancer Genome Atlas (TCGA)~\citep{tomczak2015cancer}.

\subsection{Synthetic Experiments}\label{subsec:synthetic}

\subsubsection{Data Generation}
\begin{figure}[t]
  \centering
  \vspace{.3in}
  \includegraphics[width=0.85\linewidth]{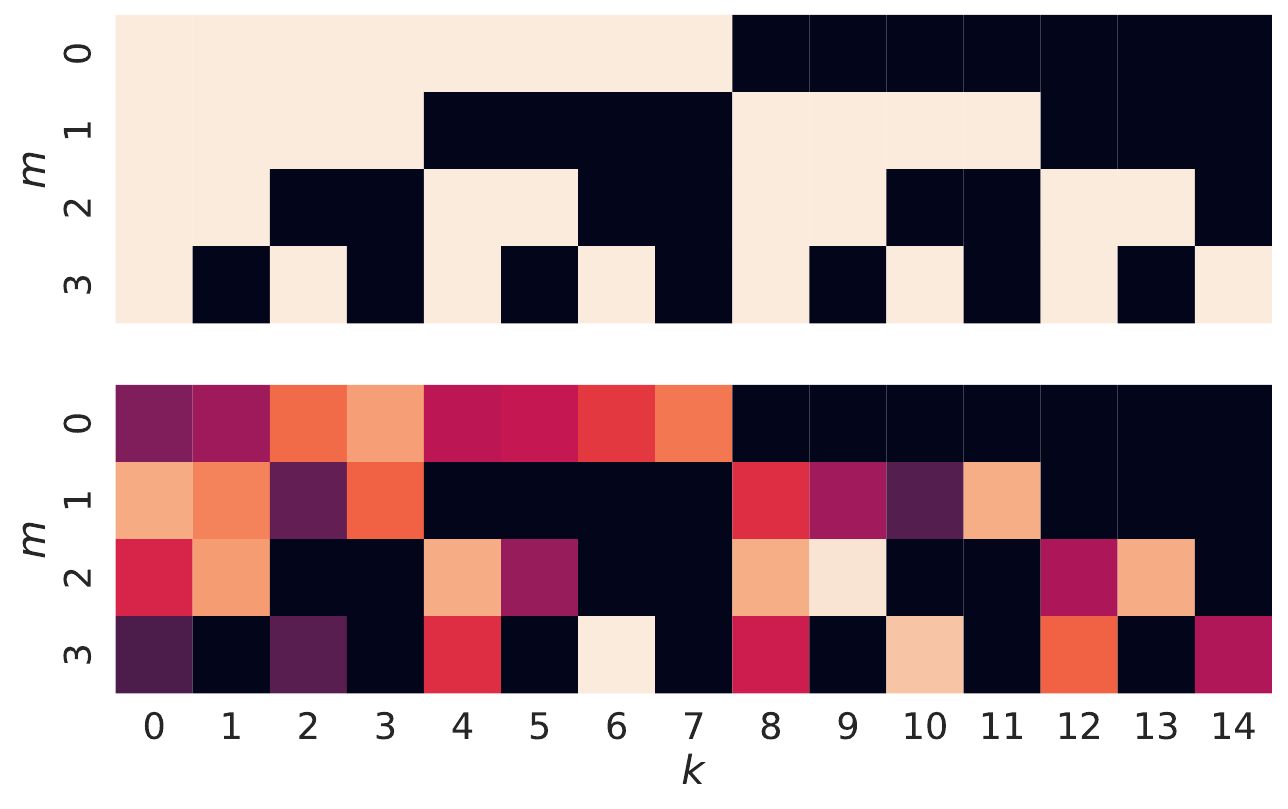}
  \vspace{.1in}
  \caption{Depiction of cross-view relationships in the synthetic data. True relationship (top) versus inferred relationship based on the factor scales learned by MuVI (bottom).}\label{fig:view_factor_scales}
\end{figure}

We adopt the data generation process from~\citep{zhao2016bayesian}, and compile a synthetic dataset of $N=200$ samples across four views, each comprising $D_1 = D_2 = D_3 = D_4 = 400$ features. The latent space consists of $K=15$ factors, that are linearly transformed by a set of sparse factor loadings. Each weight is sampled independently from a standard normal distribution, where loadings with an absolute value of less than $0.1$ are set to zero, to emphasize the gap between active and inactive signals. In addition, we randomly set $85\%$-$95\%$ of the loadings to zero. The relationship between the data views is explained by the structured sparsity of the loadings. In particular, we generate all possible combinations of four binary variables, resulting in $15$ distinct relationship configurations for each factor as depicted in the top heatmap of Figure~\ref{fig:view_factor_scales}. A dark entry means the factor does not contribute to the corresponding view, and decouples it from the rest of the views. For instance, the factor with index 0 is fully shared across all views, whereas the factors 7, 11, 13 and 14 explain variability pertaining to each view individually. The rest of the factors exhibit a mixture of partially shared configurations.
We extract the true feature sets, indicating active and inactive features for each latent dimension, and generate potentially noisy feature sets which serve as prior information during training. We perturb the feature sets by swapping a fraction of the true positives with true negatives. The severity of the noise translates to a poorly specified prior belief, where $100\%$ noise means that the prior belief is entirely incorrect, up to the number of the expected active features. Also, we introduce additional false positive features for factors that are completely inactive, as to further disguise any true underlying relationship within and between views.

\subsubsection{Evaluation}\label{subsubsec:eval}
We evaluate all models on five randomly generated datasets as described above, and report average metric scores across all views. We assess the quality of the representation in terms of the RMSE between the true observations and the reconstructed views from the common latent space. We quantify the concordance between the inferred factor loadings and the true underlying structure by reporting the precision, recall and the $F_1$ score. As a prerequisite to calculating the binary scores we extract a binary representation of active and inactive features based on a threshold. For our model and MOFA, the threshold of $0.1$ matches the true cutoff between active and inactive signals when generating the sparse loadings. For BASS and AE, the optimal thresholds were $0.05$ and $0.01$, respectively. In addition, we show the precision and recall curves summarizing results across all possible thresholds in the Appendix~\ref{appendix_a}. Finally, we perform factor matching or label switching in a post hoc manner, following~\citep{zhao2016bayesian}, where we reorder the factors to match the true order for a direct comparison with the ground truth. 

\subsubsection{Training}
The parameter and training settings for the baselines are summarized in the Appendix~\ref{appendix_a}.
We perform a simple search on the single hyperparameter of MuVI, the auxiliary constant $\alpha_{j, k}$ for the factor loadings that are not part of the prior feature sets. Among $\left\{0.01, 0.03, 0.05, 0.1\right\}$, $\alpha_{j, k} = 0.03$ reliably performs the best. A complete sensitivity analysis provided in the Appendix~\ref{appendix_b} shows that our approach is robust against the choice of $\alpha$. We train several models with different noise configuration and prior information availability. We start with an uninformed model and iteratively increase the amount of the prior information by informing a single view, two views and eventually three views. An important consideration when injecting prior information for a subset of views is assuming that the features of the rest of the views are not present in the prior feature sets, and informing them accordingly. In other words, we choose the same $\alpha_{j, k}$ for all the factor loadings for the uninformed views that we choose for the factor loadings of the informed views that are absent in the prior feature sets. The main rationale behind this decision is encouraging the informed views to learn first, and gradually inform the rest of the uninformed views implicitly via the structure of the shared factors. Also, we observed empirically that the uninformed views are unrestricted in terms of learning and typically converge very early during training and are unable to ``unlearn'' a suboptimal structure of the factor loadings.

% Please add the following required packages to your document preamble:
% \usepackage{booktabs}
% \usepackage{graphicx}
\begin{table}[t]
\caption{Performance comparison on the synthetic data. Average metric scores across five independent runs $\pm$ standard deviation. } \label{tab:main_results}
\begin{center}
\resizebox{\linewidth}{!}{%
\begin{tabular}{@{}lcccc@{}}
\toprule
\multicolumn{1}{c}{} & RMSE           & Precision      & Recall         & $F_1$       \\ \midrule\midrule
GFA   & $0.324 \pm 0.00$          & $0.256 \pm 0.05$          & $0.916 \pm 0.05$          & $0.402 \pm 0.06$          \\ \midrule
BASS   & $0.543 \pm 0.02$          & $\mathbf{0.944 \pm 0.05}$          & $0.899 \pm 0.03$          & $0.920 \pm 0.04$          \\ \midrule
MOFA               & $0.331 \pm 0.00$          & $0.541 \pm 0.14$          & $0.914 \pm 0.02$          & $0.672 \pm 0.11$          \\ \midrule
AE                   & $0.325 \pm 0.01$ & $0.692 \pm 0.06$          & $0.964 \pm 0.01$          & $0.805 \pm 0.04$          \\ \midrule
MuVI$_U$ (uninformed)    & $\mathbf{0.323 \pm 0.00}$ & $0.933 \pm 0.05$          & $\mathbf{0.995 \pm 0.00}$          & $\mathbf{0.962 \pm 0.03}$          \\ \midrule\midrule
% \multicolumn{5}{c}{50\% noise}                                                           \\ \midrule
MuVI$_{0.5}$ (1 inf. view)        & $\mathbf{0.322 \pm 0.00}$          & $0.966 \pm 0.01$          & $0.996 \pm 0.00$          & $0.981 \pm 0.00$          \\ \midrule
% \multicolumn{5}{c}{10\% noise}                                                           \\ \midrule
MuVI$_{0.1}$ (1 inf. view)        & $\mathbf{0.322 \pm 0.00}$          & $0.965 \pm 0.01$          & $0.997 \pm 0.00$ & $0.982 \pm 0.00$          \\ \midrule
MuVI$_{0.1}$ (3 inf. views)       & $\mathbf{0.322 \pm 0.00}$          & $\mathbf{0.974 \pm 0.01}$ & $\mathbf{0.998 \pm 0.00}$          & $\mathbf{0.986 \pm 0.01}$ \\ \bottomrule
\end{tabular}%
}
\end{center}
\end{table}

\subsubsection{Results}\label{subsubsec:res}
\begin{figure*}[t]
  \centering
  \vspace{.1in}
  \includegraphics[width=1.0\linewidth]{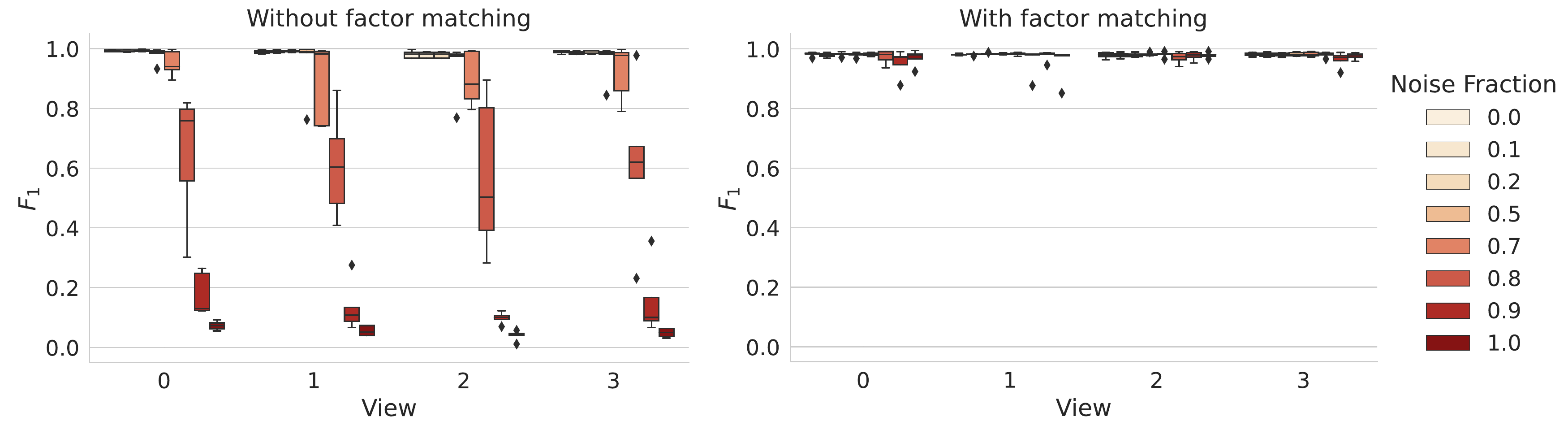}
  \vspace{.1in}
  \caption{Performance across multiple prior noise configurations when informing only a single view (view 0). Informing a factor in one view is sufficient to promote identifiability in uninformed views even for medium-to-high noise (left). Factor matching is required only for noise levels greater than $70\%$ (right). Each boxplot comprises five independent runs of the same noise fraction.}\label{fig:boxplot_noise_new}
\end{figure*}
\paragraph{Overall performance against baselines}
We provide a visual summary of the experiments in Figure~\ref{fig:teaser_new}, and a detailed overview of the results in Table~\ref{tab:main_results}. In terms of the quality of the latent representation, measured by the RMSE, most models perform comparably well. 
It is worth noting that the hybrid AE, combining a non-linear encoder with a linear decoder, does not yield a better latent representation. Recent theoretical work on posterior collapse in linear VAEs~\citep{lucas2019don} investigates and proves that there is no benefit in using a non-linear encoder when the decoder is linear. On the other hand, BASS appears to sacrifice reconstruction quality in favor of additional sparsity, which is supported by a high precision and relatively lower recall. As expected, group factor analysis (GFA) is unable to provide sparse solutions. Our modeling approach strikes a better balance between learning a good latent representation with low reconstruction error and pruning away the majority of superfluous connections between the latent factors and the observed features. In addition, the high recall indicates a better capability of MuVI$_U$ to retrieve virtually all positive signals, which results in the highest $F_1$ score, even in the absence of an informative prior.
% outperforms state-of-the-art approaches in terms of the reconstruction error and precision, recall and $F_1$ score 
In addition to the uninformed version of our model (MuVI$_U$), we report the results for the informed models. For the first model (MuVI$_{0.5}$), we simulate a realistic scenario where the available information is very limited and noisy. Specifically, we inform only the first view with highly perturbed feature sets with a noise fraction of $50\%$ (MuVI$_{0.5}$). Compared to the uninformed model, we notice an increase in the precision, while the recall remains intact, thereby further improving the $F_1$ score. In the next two models, we first reduce the amount of noise to $10\%$ (MuVI$_{0.1}$), then increase the number of informed views from one to three. We observe additional improvements in all three binary scores, which further emphasizes the benefit of increasing the quality and the quantity of the prior information. 

\paragraph{Learning from noisy priors}
Next, we assess the robustness of MuVI against severe noise, and its capability of utilizing and transferring useful prior information across all views. Figure~\ref{fig:boxplot_noise_new} depicts a comprehensive overview of the results when informing only a single view, while increasing the amount of prior noise. We first assess the extent to which MuVI is able to infer identifiable factors for increasing levels of noise. Fig.~\ref{fig:boxplot_noise_new} (left) shows that up to a noise fraction of around $70\%$, MuVI exhibits no difficulties in learning a good representation, and identifies the true underlying structure of the latent factors with a median $F_1$ score of almost 1.0, not only for the informed view, but also for the uninformed views. This indicates that MuVI is able to promote identifiability across views.
Recovering the true structure of the latent factors first becomes challenging when increasing the noise fraction beyond $80\%$, with a median $F_1$ score of $0.62$ for $80\%$ noise. As expected, a severe perturbation of at least $90\%$ significantly reduces the ability of our model to find the proper ordering of the factors. However, the plot on the right shows that the model merely loses its ability to identify the correct order of the latent dimensions, as the corrected permutation of the inferred factors corresponds to the true structure across all factors and data views. In the Appendix~\ref{appendix_b} we provide additional results by considering the noise fraction as a proxy for the number of pure variables available in the prior information.\\
Furthermore, the quality of the representation remains virtually intact. For comparison, the mean RMSE for the trained MuVI models with a noise fraction of $80\%$ or less is $0.322$ for each noise level, and for a noise fraction of $90\%$ and $100\%$ we observe only a marginal increase of the RMSE to $0.326$ and $0.328$, respectively (see Appendix~\ref{appendix_b}). These results still outperform the baselines, and further assert that any amount of prior information benefits our model highly, while a (fully) incorrect prior information does not harm its performance.
\vspace{-3mm}
\paragraph{Learning cross-view relationships}
Finally, we look at the ability of MuVI to communicate the underlying relationship between the views. The bottom heatmap in Figure~\ref{fig:view_factor_scales} corresponds to the MAP estimates of the factor scales $\delta_{k}^{(m)}$ learned during inference. The model successfully learns to turn off irrelevant factors and highlights cross-view relationships by assigning large positive values to active and shared factors.
% , thereby diminishing the regularization penalty imposed by the sparse prior.

\subsection{The Cancer Genome Atlas (TCGA)}\label{subsec:tcga}

\begin{figure*}[t]
  \centering
  \vspace{.1in}
  \includegraphics[width=1.0\linewidth]{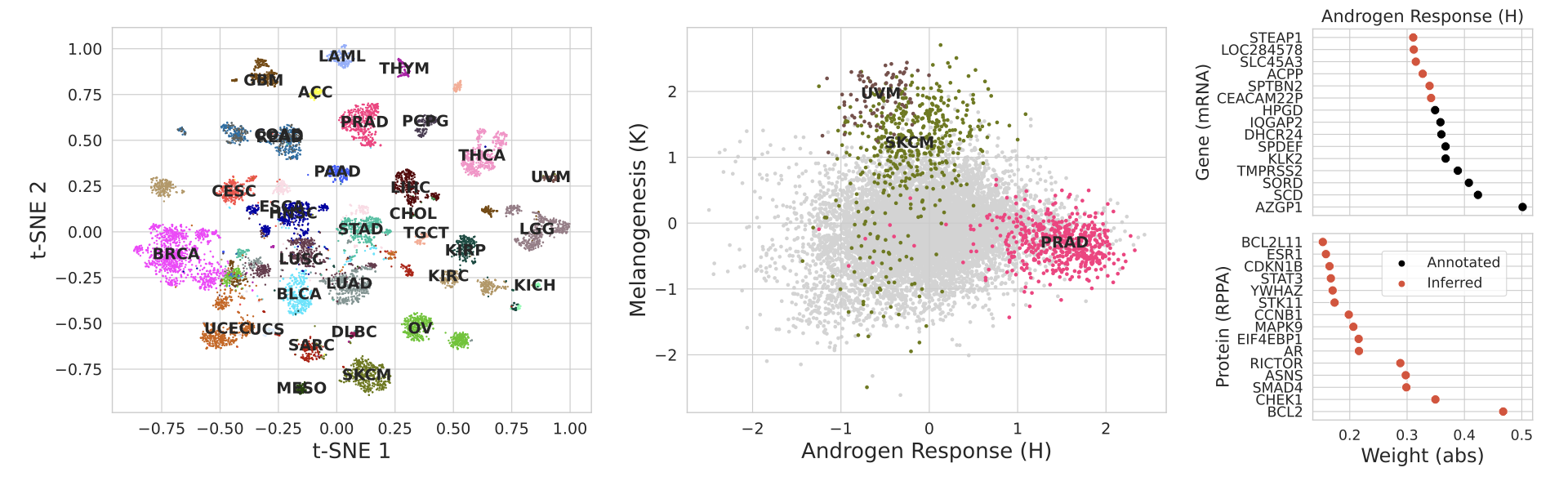}
  % \vspace{.1in}
  \caption{Results of the TCGA dataset. A t-SNE embedding space learned from the latent representation of MuVI (left). Inferred latent factors that are informed by androgen response and melanogenesis a priori (center). Top 15 features based on the absolute value of their corresponding factor loadings for the shared androgen response factor among mRNA and RPPA (right).}\label{fig:tcga_main}
\end{figure*}

We investigate a large dataset of comprehensive multi-omic profiling of over 11 thousand samples from 33 cancer types~\citep{tomczak2015cancer}. Each sample comprises sets of features of various sizes across four biologically distinct views: DNA methylation (6,000), mRNA expression (6,000), microRNA expression (728) and reverse phase protein array or RPPA (312). We include a preprocessing step of centering the data and standardizing each view globally due to the large differences in value ranges across views. In addition, samples exhibit partial or complete missingness in individual views, which we accommodate in our probabilistic modeling approach.
Prior to training, we derive gene set annotations from curated public databases such as MSigDB~\citep{liberzon2015molecular}, Reactome~\citep{fabregat2018reactome} and KEGG~\citep{kanehisa2000kegg}. A gene set consists of a group of genes that are biologically meaningful when co-expressed under certain conditions, or describe functionally distinct pathways in biological systems. Since the observed features are genes, each gene set acts as a prior feature set for the factor loadings, thereby encoding a well-defined biological pathway into the corresponding latent factor.
As informative prior we consider gene set annotations where at least 15 genes were present in the data, resulting in 360 annotations with a median size of 30. Since gene sets encompass only genomic and transcriptomic features, we are limited to informing only the first two views, namely the DNA and the mRNA. 
We take advantage of the stochastic variational inference algorithm and propagate batches of 1,000 samples during training. The algorithm terminates when the optimization objective no longer improves significantly in consecutive iterations, after a certain number of patience steps. The training converges in less than 12 minutes on a single NVIDIA Quadro RTX 5000 GPU with 16 GB memory.
% \newpage
\paragraph{Global structure}
First, we assess the quality of the latent space inferred by our model. 
We systematically evaluate and compare the performance of MuVI and the other baselines in terms of explaining the underlying structure of the data introduced by the different cancer types.
For each model, we first infer and extract the latent representation of each observation. Next, we apply a t-distributed stochastic neighbor embedding (t-SNE) approach to further compress the latent space into two dimensions for better visualization~\citep{van2008visualizing}. 
In the leftmost plot of Figure~\ref{fig:tcga_main} we display the new embeddings. Our model is able to learn a meaningful structure by grouping samples of a similar type closer together. Subsequently, we apply a K-Means algorithm by fixing the number of clusters to the number of known cancer types (33). Finally, we compare the learned clusters by the K-Means with the true underlying cancer types and compute the cluster purity and the adjusted Rand index (ARI) to measure the similarity between both clusterings. We perform the same procedure several times and report the average scores in the first two columns of Table~\ref{tab:tcga_ari}.
The scores reiterate the results obtained in the synthetic evaluation. Overall, MOFA, AE and MuVI perform comparably well by capturing known cancer types. On the other hand, BASS infers a less informative latent space as indicated by the multiple cluster overlaps (see Appendix~\ref{appendix_c}), and the lower clustering scores. 
Importantly, MuVI is inherently interpretable with each learned factor being directly labeled by a biological pathway: our model intrinsically accommodates prior information in terms of feature sets, thereby guiding the inference of pre-tagged axes of variation from domain expertise. In contrast, rendering the baseline methods interpretable requires a cumbersome manual inspection of the loadings of each factor, with subsequent enrichment analysis searching for pathways in which large loadings are over-represented.

\paragraph{Relevant factors}
Next, we apply a one-vs-rest Wilcoxon rank test to identify the main pathways that drive the heterogeneity of patients with respect to their cancer types. A complete overview of the test results is provided in the Appendix~\ref{appendix_c}. 
In particular, the results hint at melanogenesis (KEGG) and the androgen response (Hallmark) as relevant pathways for melanoma patients and prostate adenocarcinoma patients, respectively.
Melanogenesis and the androgen response describe general biological processes. Melanogenesis describes the process of producing the melanin pigments from melanocytes, commonly found in the epidermis and hair follicles. Androgen response orchestrates the activation of male hormone receptors, and plays an important role in the development and progression of prostate cancer~\citep{fujita2019role}. In the middle of Figure~\ref{fig:tcga_main} we illustrate a scatter plot by mapping our samples onto the latent space spanned by the two axes informed and described by androgen response and melanogenesis. We highlight three cancer types with the same coloring as the previous t-SNE plot, while fading the remaining 30 cancer types to gray. As anticipated, melanogenesis helps identify the only two groups of samples from tissues with skin cutaneous melanoma (SKCM) and uveal melanoma (UVM). Likewise, androgen response assigns significantly higher scores to patients with prostate adenocarcinoma (PRAD). We assess how predictive each latent factor is to its corresponding cancer type by reporting the average area under the ROC curve (AUC) computed for the latent factors assigned to each cancer type in a one-vs-rest fashion. We compare the obtained results with the baselines, by performing a gene set enrichment analysis~\citep{subramanian2005gene} on the factor loadings inferred by BASS, MOFA and AE, and match each latent factor to a gene set.
Then, we focus on the latent factors that model androgen response and melanogenesis, and measure the one-vs-rest AUC of the factor scores for PRAD and SKCM + UVM, respectively. Since the same pathway can be assigned to multiple latent factors for MOFA and AE, we select the best performing factor and report the AUC scores in the last two columns of Table~\ref{tab:tcga_ari}.
MuVI achieves the highest AUC, while rendering each latent factor directly interpretable due to the encoded pathway knowledge a priori.
% thereby rendering each latent factor inherently interpretable.

\begin{table}[t]
\caption{Performance comparison on the TCGA dataset. Summary of the cluster purity and the adjusted Rand index (ARI) for measuring the similarity between the known cancer subtypes and the learned clusters by the K-Means algorithm (first two columns). In addition, we report the average AUC scores of individual factors modeling androgen response and melanogenesis (last two columns). For the baselines it is necessary to perform an additional gene set enrichment analysis (GSEA) to label each factor.} \label{tab:tcga_ari}
\begin{center}
\vspace{.2in}
\resizebox{\linewidth}{!}{%
\begin{tabular}{@{}lcccc@{}}
\toprule
 & ARI & Purity & \begin{tabular}[c]{@{}c@{}}Androgen \\ Resp. (AUC)\end{tabular} & \begin{tabular}[c]{@{}c@{}}Melanogenesis \\ (AUC)\end{tabular}\\ \midrule
BASS & $0.609\pm 0.01$ & $0.763\pm 0.01$ & $0.553 \pm 0.04$ & $0.531 \pm 0.01$ \\ \midrule
MOFA & $0.658\pm 0.01$ & $0.822\pm 0.01$ & $0.569 \pm 0.04$ & $0.528 \pm 0.01$ \\ \midrule
AE & $\mathbf{0.662\pm 0.01}$ & $\mathbf{0.828\pm 0.01}$ & $0.821 \pm 0.05$ & $0.724 \pm 0.02$ \\ \midrule
MuVI & $0.660\pm0.01$ & $0.825\pm 0.01$ & $\mathbf{0.910 \pm 0.03}$ & $\mathbf{0.872 \pm 0.02}$ \\ \bottomrule
\end{tabular}
}%
\end{center}
\end{table}

\paragraph{Factor inspection}
Finally, we focus on the composition of the inferred latent factors. In the rightmost plots, we inspect the top 15 features of the androgen response factor, which is active and shared among the mRNA and the RPPA view. We decide whether a factor is active in view $m$ if its corresponding factor scale is not virtually zero, i.e.\ $\delta_{k}^{(m)}>0.01$, as well as if the variance explained by the factor exceeds a threshold of $R^2 > 0.5\%$. We reiterate that we informed only mRNA among the two views. Hence, we expect to see annotated genes weighted highly based on the absolute value of their corresponding factor loadings. 
A powerful feature of MuVI is the refinement of provided feature sets by adding or removing features in a data-driven manner. Investigating inferred genes for the androgen response factor, we discover several previously known biomarkers of clinical relevance such as STEAP1~\citep{khanna2021clinical}, SLC45A3~\citep{perner2013loss} and ACPP~\citep{kong2013emerging}, which are overexpressed in patients with prostate cancer.
% A powerful feature of MuVI is the refinement of provided feature sets by adding or removing features in a data-driven manner. Investigating inferred genes for the androgen response factor, we discover several previously known biomarkers of clinical relevance such as STEAP1, SLC45A3\citep{perner2013loss} and ACPP. The six-transmembrane epithelial antigen of the prostate 1 (STEAP1) is overexpressed in prostate cancer and serves as a diagnostic and prognostic biomarker\citep{khanna2021clinical}. Also, it has been shown that levels of ACPP or prostatic specific acid phosphatase (PSAP), increase with prostate cancer progression\citep{kong2013emerging}.
We now move to the uninformed view, where we expect the androgen response factor of mRNA to influence the structure of the factor loadings in the RPPA, such that the inferred factor reveals protein markers linked to prostate cancer. 
% Based on the amount of variance explained by this factor in the RPPA view, and the significance of the shared and informed factor from the mRNA view, we have strong evidence about the association of the top proteins with the disease. 
Indeed, the inferred BCL2 is an oncoprotein that inhibits the process of apoptosis~\citep{lin2007up}, whereas targeting the checkpoint kinase 1 (CHEK1) leads to favorable clinical outcomes for prostate cancer patients~\citep{drapela2020chk1}. 
% It has been demonstrated that BCL2 is a prerequisite for the progression of cancer cells from an androgen-dependent stage, the earlier stages of the cancer progression, to an androgen-independent growth stage~\citep{lin2007up}. 

\section{CONCLUSION}\label{sec:conc}
In this contribution, we have addressed the task of modeling multi-view data with interpretable latent variable models. We use a Bayesian approach with structured sparsity to encode domain knowledge in the priors of a factor analysis model. Our model is able to integrate noisy domain expertise and results in factors that are inherently interpretable via their pre-defined feature sets. In contrast to other sparse multi-view models, our approach is able to recover correct signals from noisy feature sets, while maintaining a low reconstruction error. We demonstrate in a real-world application that our model is able to infer biologically meaningful and clinically relevant subpopulations of cancer patients. Moreover, the ability of MuVI to transfer prior knowledge from informed to uninformed views via the shared factors shows the potential of discovering novel feature sets in related domains. %Partially informed novel feature sets discovered by our have the potential to improve treatment therapies.

\bibliography{main}

\begin{thebibliography}{48}
\providecommand{\natexlab}[1]{#1}
\providecommand{\url}[1]{\texttt{#1}}
\expandafter\ifx\csname urlstyle\endcsname\relax
  \providecommand{\doi}[1]{doi: #1}\else
  \providecommand{\doi}{doi: \begingroup \urlstyle{rm}\Url}\fi

\bibitem[Ainsworth et~al.(2018)Ainsworth, Foti, Lee, and Fox]{ainsworth2018oi}
Samuel~K Ainsworth, Nicholas~J Foti, Adrian~KC Lee, and Emily~B Fox.
\newblock oi-vae: Output interpretable vaes for nonlinear group factor
  analysis.
\newblock In \emph{International Conference on Machine Learning}, pages
  119--128. PMLR, 2018.

\bibitem[Anderson and Rubin(1956)]{anderson1956statistical}
TW~Anderson and Herman Rubin.
\newblock Statistical inference in.
\newblock In \emph{Proceedings of the Third Berkeley Symposium on Mathematical
  Statistics and Probability: Held at the Statistical Laboratory, University of
  California, December, 1954, July and August, 1955}, volume~1, page 111. Univ
  of California Press, 1956.

\bibitem[Argelaguet et~al.(2018)Argelaguet, Velten, Arnol, Dietrich, Zenz,
  Marioni, Buettner, Huber, and Stegle]{argelaguet2018multi}
Ricard Argelaguet, Britta Velten, Damien Arnol, Sascha Dietrich, Thorsten Zenz,
  John~C Marioni, Florian Buettner, Wolfgang Huber, and Oliver Stegle.
\newblock Multi-omics factor analysis—a framework for unsupervised
  integration of multi-omics data sets.
\newblock \emph{Molecular systems biology}, 14\penalty0 (6):\penalty0 e8124,
  2018.

\bibitem[Armagan et~al.(2011)Armagan, Clyde, and
  Dunson]{armagan2011generalized}
Artin Armagan, Merlise Clyde, and David Dunson.
\newblock Generalized beta mixtures of gaussians.
\newblock \emph{Advances in neural information processing systems}, 24, 2011.

\bibitem[Barbie et~al.(2009)Barbie, Tamayo, Boehm, Kim, Moody, Dunn, Schinzel,
  Sandy, Meylan, Scholl, et~al.]{barbie2009systematic}
David~A Barbie, Pablo Tamayo, Jesse~S Boehm, So~Young Kim, Susan~E Moody, Ian~F
  Dunn, Anna~C Schinzel, Peter Sandy, Etienne Meylan, Claudia Scholl, et~al.
\newblock Systematic rna interference reveals that oncogenic kras-driven
  cancers require tbk1.
\newblock \emph{Nature}, 462\penalty0 (7269):\penalty0 108--112, 2009.

\bibitem[Bernardo et~al.(2003)Bernardo, Bayarri, Berger, Dawid, Heckerman,
  Smith, and West]{bernardo2003bayesian}
JM~Bernardo, MJ~Bayarri, JO~Berger, AP~Dawid, D~Heckerman, AFM Smith, and
  M~West.
\newblock Bayesian factor regression models in the “large p, small n”
  paradigm.
\newblock \emph{Bayesian statistics}, 7:\penalty0 733--742, 2003.

\bibitem[Bing et~al.(2020)Bing, Bunea, Ning, and Wegkamp]{bing2020adaptive}
Xin Bing, Florentina Bunea, Yang Ning, and Marten Wegkamp.
\newblock Adaptive estimation in structured factor models with applications to
  overlapping clustering.
\newblock \emph{The Annals of Statistics}, 48\penalty0 (4):\penalty0
  2055--2081, 2020.

\bibitem[Bingham et~al.(2019)Bingham, Chen, Jankowiak, Obermeyer, Pradhan,
  Karaletsos, Singh, Szerlip, Horsfall, and Goodman]{bingham2019pyro}
Eli Bingham, Jonathan~P Chen, Martin Jankowiak, Fritz Obermeyer, Neeraj
  Pradhan, Theofanis Karaletsos, Rohit Singh, Paul Szerlip, Paul Horsfall, and
  Noah~D Goodman.
\newblock Pyro: Deep universal probabilistic programming.
\newblock \emph{The Journal of Machine Learning Research}, 20\penalty0
  (1):\penalty0 973--978, 2019.

\bibitem[Buettner et~al.(2017)Buettner, Pratanwanich, McCarthy, Marioni, and
  Stegle]{buettner2017f}
Florian Buettner, Naruemon Pratanwanich, Davis~J McCarthy, John~C Marioni, and
  Oliver Stegle.
\newblock f-sclvm: scalable and versatile factor analysis for single-cell
  rna-seq.
\newblock \emph{Genome biology}, 18\penalty0 (1):\penalty0 1--13, 2017.

\bibitem[Cao and Fleet(2014)]{cao2014generalized}
Yanshuai Cao and David~J Fleet.
\newblock Generalized product of experts for automatic and principled fusion of
  gaussian process predictions.
\newblock \emph{arXiv preprint arXiv:1410.7827}, 2014.

\bibitem[Carvalho et~al.(2009)Carvalho, Polson, and
  Scott]{carvalho2009handling}
Carlos~M Carvalho, Nicholas~G Polson, and James~G Scott.
\newblock Handling sparsity via the horseshoe.
\newblock In \emph{Artificial Intelligence and Statistics}, pages 73--80. PMLR,
  2009.

\bibitem[Carvalho et~al.(2010)Carvalho, Polson, and
  Scott]{carvalho2010horseshoe}
Carlos~M Carvalho, Nicholas~G Polson, and James~G Scott.
\newblock The horseshoe estimator for sparse signals.
\newblock \emph{Biometrika}, 97\penalty0 (2):\penalty0 465--480, 2010.

\bibitem[Damianou et~al.(2012)Damianou, Ek, Titsias, and
  Lawrence]{damianou2012manifold}
AC~Damianou, Carl~Henrik Ek, MK~Titsias, and ND~Lawrence.
\newblock Manifold relevance determination.
\newblock In \emph{29th International Conference on Machine Learning, ICML
  2012, 26 June 2012 through 1 July 2012, Edinburgh}, pages 145--152, 2012.

\bibitem[Dr{\'a}pela et~al.(2020)Dr{\'a}pela, Khirsariya, van Weerden, Fedr,
  Such{\'a}nkov{\'a}, B{\'u}zov{\'a}, {\v{C}}erven{\`y}, Hampl, Puhr, Watson,
  et~al.]{drapela2020chk1}
Stanislav Dr{\'a}pela, Prashant Khirsariya, Wytske~M van Weerden, Radek Fedr,
  Tereza Such{\'a}nkov{\'a}, Diana B{\'u}zov{\'a}, Jan {\v{C}}erven{\`y},
  Ale{\v{s}} Hampl, Martin Puhr, William~R Watson, et~al.
\newblock The chk1 inhibitor mu380 significantly increases the sensitivity of
  human docetaxel-resistant prostate cancer cells to gemcitabine through the
  induction of mitotic catastrophe.
\newblock \emph{Molecular oncology}, 14\penalty0 (10):\penalty0 2487--2503,
  2020.

\bibitem[Engelhardt and Stephens(2010)]{engelhardt2010analysis}
Barbara~E Engelhardt and Matthew Stephens.
\newblock Analysis of population structure: a unifying framework and novel
  methods based on sparse factor analysis.
\newblock \emph{PLoS genetics}, 6\penalty0 (9):\penalty0 e1001117, 2010.

\bibitem[Fabregat et~al.(2018)Fabregat, Jupe, Matthews, Sidiropoulos,
  Gillespie, Garapati, Haw, Jassal, Korninger, May,
  et~al.]{fabregat2018reactome}
Antonio Fabregat, Steven Jupe, Lisa Matthews, Konstantinos Sidiropoulos, Marc
  Gillespie, Phani Garapati, Robin Haw, Bijay Jassal, Florian Korninger, Bruce
  May, et~al.
\newblock The reactome pathway knowledgebase.
\newblock \emph{Nucleic acids research}, 46\penalty0 (D1):\penalty0 D649--D655,
  2018.

\bibitem[Frost et~al.(2015)Frost, Li, and Moore]{frost2015principal}
H~Robert Frost, Zhigang Li, and Jason~H Moore.
\newblock Principal component gene set enrichment (pcgse).
\newblock \emph{BioData mining}, 8\penalty0 (1):\penalty0 1--18, 2015.

\bibitem[Fujita and Nonomura(2019)]{fujita2019role}
Kazutoshi Fujita and Norio Nonomura.
\newblock Role of androgen receptor in prostate cancer: a review.
\newblock \emph{The world journal of men's health}, 37\penalty0 (3):\penalty0
  288--295, 2019.

\bibitem[Ghosh et~al.(2018)Ghosh, Yao, and Doshi-Velez]{ghosh2018structured}
Soumya Ghosh, Jiayu Yao, and Finale Doshi-Velez.
\newblock Structured variational learning of bayesian neural networks with
  horseshoe priors.
\newblock In \emph{International Conference on Machine Learning}, pages
  1744--1753. PMLR, 2018.

\bibitem[Hoffman et~al.(2013)Hoffman, Blei, Wang, and
  Paisley]{hoffman2013stochastic}
Matthew~D Hoffman, David~M Blei, Chong Wang, and John Paisley.
\newblock Stochastic variational inference.
\newblock \emph{Journal of Machine Learning Research}, 2013.

\bibitem[Hotelling(1992)]{hotelling1992relations}
Harold Hotelling.
\newblock Relations between two sets of variates.
\newblock In \emph{Breakthroughs in statistics}, pages 162--190. Springer,
  1992.

\bibitem[Kanehisa and Goto(2000)]{kanehisa2000kegg}
Minoru Kanehisa and Susumu Goto.
\newblock Kegg: kyoto encyclopedia of genes and genomes.
\newblock \emph{Nucleic acids research}, 28\penalty0 (1):\penalty0 27--30,
  2000.

\bibitem[Khanna et~al.(2021)Khanna, Salmond, Lynn, Leong, and
  Williams]{khanna2021clinical}
Karan Khanna, Nikki Salmond, Kalan~S Lynn, Hon~S Leong, and Karla~C Williams.
\newblock Clinical significance of steap1 extracellular vesicles in prostate
  cancer.
\newblock \emph{Prostate cancer and prostatic diseases}, 24\penalty0
  (3):\penalty0 802--811, 2021.

\bibitem[Kingma and Welling(2013)]{kingma2013auto}
Diederik~P Kingma and Max Welling.
\newblock Auto-encoding variational bayes.
\newblock \emph{arXiv preprint arXiv:1312.6114}, 2013.

\bibitem[Klami et~al.(2013)Klami, Virtanen, and Kaski]{klami2013bayesian}
Arto Klami, Seppo Virtanen, and Samuel Kaski.
\newblock Bayesian canonical correlation analysis.
\newblock \emph{Journal of Machine Learning Research}, 14\penalty0 (4), 2013.

\bibitem[Klami et~al.(2014)Klami, Virtanen, Lepp{\"a}aho, and
  Kaski]{klami2014group}
Arto Klami, Seppo Virtanen, Eemeli Lepp{\"a}aho, and Samuel Kaski.
\newblock Group factor analysis.
\newblock \emph{IEEE transactions on neural networks and learning systems},
  26\penalty0 (9):\penalty0 2136--2147, 2014.

\bibitem[Kong and Byun(2013)]{kong2013emerging}
Hoon~Young Kong and Jonghoe Byun.
\newblock Emerging roles of human prostatic acid phosphatase.
\newblock \emph{Biomolecules \& therapeutics}, 21\penalty0 (1):\penalty0 10,
  2013.

\bibitem[Lan et~al.(2014)Lan, Waters, Studer, and Baraniuk]{lan2014sparse}
Andrew~S Lan, Andrew~E Waters, Christoph Studer, and Richard~G Baraniuk.
\newblock Sparse factor analysis for learning and content analytics.
\newblock \emph{The Journal of Machine Learning Research}, 15\penalty0
  (1):\penalty0 1959--2008, 2014.

\bibitem[Lee and van~der Schaar(2021)]{lee2021variational}
Changhee Lee and Mihaela van~der Schaar.
\newblock A variational information bottleneck approach to multi-omics data
  integration.
\newblock In \emph{International Conference on Artificial Intelligence and
  Statistics}, pages 1513--1521. PMLR, 2021.

\bibitem[Liberzon et~al.(2015)Liberzon, Birger, Thorvaldsd{\'o}ttir, Ghandi,
  Mesirov, and Tamayo]{liberzon2015molecular}
Arthur Liberzon, Chet Birger, Helga Thorvaldsd{\'o}ttir, Mahmoud Ghandi, Jill~P
  Mesirov, and Pablo Tamayo.
\newblock The molecular signatures database hallmark gene set collection.
\newblock \emph{Cell systems}, 1\penalty0 (6):\penalty0 417--425, 2015.

\bibitem[Lin et~al.(2007)Lin, Fukuchi, Hiipakka, Kokontis, and
  Xiang]{lin2007up}
Yuting Lin, Junichi Fukuchi, Richard~A Hiipakka, John~M Kokontis, and Jialing
  Xiang.
\newblock Up-regulation of bcl-2 is required for the progression of prostate
  cancer cells from an androgen-dependent to an androgen-independent growth
  stage.
\newblock \emph{Cell research}, 17\penalty0 (6):\penalty0 531--536, 2007.

\bibitem[Lotfollahi et~al.(2023)Lotfollahi, Rybakov, Hrovatin, Hediyeh-Zadeh,
  Talavera-L{\'o}pez, Misharin, and Theis]{lotfollahi2023biologically}
Mohammad Lotfollahi, Sergei Rybakov, Karin Hrovatin, Soroor Hediyeh-Zadeh,
  Carlos Talavera-L{\'o}pez, Alexander~V Misharin, and Fabian~J Theis.
\newblock Biologically informed deep learning to query gene programs in
  single-cell atlases.
\newblock \emph{Nature Cell Biology}, pages 1--14, 2023.

\bibitem[Lucas et~al.(2019)Lucas, Tucker, Grosse, and Norouzi]{lucas2019don}
James Lucas, George Tucker, Roger~B Grosse, and Mohammad Norouzi.
\newblock Don't blame the elbo! a linear vae perspective on posterior collapse.
\newblock \emph{Advances in Neural Information Processing Systems}, 32, 2019.

\bibitem[MacKay et~al.(1994)]{mackay1994bayesian}
David~JC MacKay et~al.
\newblock Bayesian nonlinear modeling for the prediction competition.
\newblock \emph{ASHRAE transactions}, 100\penalty0 (2):\penalty0 1053--1062,
  1994.

\bibitem[Mitchell and Beauchamp(1988)]{mitchell1988bayesian}
Toby~J Mitchell and John~J Beauchamp.
\newblock Bayesian variable selection in linear regression.
\newblock \emph{Journal of the american statistical association}, 83\penalty0
  (404):\penalty0 1023--1032, 1988.

\bibitem[Park and Casella(2008)]{park2008bayesian}
Trevor Park and George Casella.
\newblock The bayesian lasso.
\newblock \emph{Journal of the American Statistical Association}, 103\penalty0
  (482):\penalty0 681--686, 2008.

\bibitem[Perner et~al.(2013)Perner, Rupp, Braun, Rubin, Moch, Dietel, Wernert,
  Jung, Stephan, and Kristiansen]{perner2013loss}
Sven Perner, Niels~J Rupp, Martin Braun, Mark~A Rubin, Holger Moch, Manfred
  Dietel, Nicolas Wernert, Klaus Jung, Carsten Stephan, and Glen Kristiansen.
\newblock Loss of slc45a3 protein (prostein) expression in prostate cancer is
  associated with slc45a3-erg gene rearrangement and an unfavorable clinical
  course.
\newblock \emph{International journal of cancer}, 132\penalty0 (4):\penalty0
  807--812, 2013.

\bibitem[Piironen and Vehtari(2017)]{piironen2017sparsity}
Juho Piironen and Aki Vehtari.
\newblock Sparsity information and regularization in the horseshoe and other
  shrinkage priors.
\newblock \emph{Electronic Journal of Statistics}, 11\penalty0 (2):\penalty0
  5018--5051, 2017.

\bibitem[Polson and Scott(2010)]{polson2010shrink}
Nicholas~G Polson and James~G Scott.
\newblock Shrink globally, act locally: Sparse bayesian regularization and
  prediction.
\newblock \emph{Bayesian statistics}, 9\penalty0 (501-538):\penalty0 105, 2010.

\bibitem[Ranganath et~al.(2014)Ranganath, Gerrish, and
  Blei]{ranganath2014black}
Rajesh Ranganath, Sean Gerrish, and David Blei.
\newblock Black box variational inference.
\newblock In \emph{Artificial intelligence and statistics}, pages 814--822.
  PMLR, 2014.

\bibitem[Ro{\v{c}}kov{\'a} and George(2018)]{rovckova2018spike}
Veronika Ro{\v{c}}kov{\'a} and Edward~I George.
\newblock The spike-and-slab lasso.
\newblock \emph{Journal of the American Statistical Association}, 113\penalty0
  (521):\penalty0 431--444, 2018.

\bibitem[Subramanian et~al.(2005)Subramanian, Tamayo, Mootha, Mukherjee, Ebert,
  Gillette, Paulovich, Pomeroy, Golub, Lander, et~al.]{subramanian2005gene}
Aravind Subramanian, Pablo Tamayo, Vamsi~K Mootha, Sayan Mukherjee, Benjamin~L
  Ebert, Michael~A Gillette, Amanda Paulovich, Scott~L Pomeroy, Todd~R Golub,
  Eric~S Lander, et~al.
\newblock Gene set enrichment analysis: a knowledge-based approach for
  interpreting genome-wide expression profiles.
\newblock \emph{Proceedings of the National Academy of Sciences}, 102\penalty0
  (43):\penalty0 15545--15550, 2005.

\bibitem[Svensson et~al.(2020)Svensson, Gayoso, Yosef, and
  Pachter]{svensson2020interpretable}
Valentine Svensson, Adam Gayoso, Nir Yosef, and Lior Pachter.
\newblock Interpretable factor models of single-cell rna-seq via variational
  autoencoders.
\newblock \emph{Bioinformatics}, 36\penalty0 (11):\penalty0 3418--3421, 2020.

\bibitem[Thurstone(1931)]{thurstone1931multiple}
Louis~Leon Thurstone.
\newblock Multiple factor analysis.
\newblock \emph{Psychological review}, 38\penalty0 (5):\penalty0 406, 1931.

\bibitem[Tibshirani(1996)]{tibshirani1996regression}
Robert Tibshirani.
\newblock Regression shrinkage and selection via the lasso.
\newblock \emph{Journal of the Royal Statistical Society: Series B
  (Methodological)}, 58\penalty0 (1):\penalty0 267--288, 1996.

\bibitem[Tomczak et~al.(2015)Tomczak, Czerwi{\'n}ska, and
  Wiznerowicz]{tomczak2015cancer}
Katarzyna Tomczak, Patrycja Czerwi{\'n}ska, and Maciej Wiznerowicz.
\newblock The cancer genome atlas (tcga): an immeasurable source of knowledge.
\newblock \emph{Contemporary oncology}, 19\penalty0 (1A):\penalty0 A68, 2015.

\bibitem[Van~der Maaten and Hinton(2008)]{van2008visualizing}
Laurens Van~der Maaten and Geoffrey Hinton.
\newblock Visualizing data using t-sne.
\newblock \emph{Journal of machine learning research}, 9\penalty0 (11), 2008.

\bibitem[Zhao et~al.(2016)Zhao, Gao, Mukherjee, and
  Engelhardt]{zhao2016bayesian}
Shiwen Zhao, Chuan Gao, Sayan Mukherjee, and Barbara~E Engelhardt.
\newblock Bayesian group factor analysis with structured sparsity.
\newblock \emph{The Journal of Machine Learning Research}, 2016.

\end{thebibliography}
\clearpage

\appendix
% % NOTE: necessary when ptmx or no mathfont class option is given
% \providecommand{\upGamma}{\Gamma}
% \providecommand{\uppi}{\pi}

\section{APPENDIX A}\label{appendix_a}
Here, we provide additional details regarding the synthetic experiments. In particular, we extend the description of the baselines and their respective training settings. We then specify the evaluation metrics such as the RMSE for measuring the quality of the reconstruction, and additional binary metrics such as precision, recall and the $F_1$ score for assessing how well each model recovers the true underlying structure of the latent factors. We briefly describe the process of factor matching applied to all models in a post-processing step. Finally, we report additional results such as the precision and recall curves for the benchmark across all possible thresholds, and the performance of MuVI when dealing with significantly higher dimensional data.
\subsection{Baselines}
The parameter and training settings for the baselines are provided here.
We train MOFA combining the ARD and the spike-and-slab prior on the weights for column-wise and element-wise sparsity. We apply BASS with default parameters and $20$ parameter-expanded expectation maximization (PX-EM) iterations.
The autoencoder consists of an encoder-decoder pair for each observed view, connected via a common latent space governed by a product of experts (PoE) approach~\citep{lee2021variational, cao2014generalized}. Each encoder comprises two hidden layers of size 64 and 32, followed by a ReLU activation. Each decoder then performs a linear mapping of the latent code to the observed features of each view. To achieve sparsity in the factor-to-feature mapping, we introduce an L1 penalty term with $\lambda=0.01$ in the respective decoder weights. All models were trained until an early stopping condition was met, that is, until the optimization objective no longer improved in consecutive iterations, and after a certain number of patience steps.

\subsection{Evaluation}
To simplify notation, we omit the view index $m$ and rewrite the collection of observations as a single matrix $\mathbf{Y} \in \R^{N\times D}$ comprising $N$ samples across $M$ coupled views $\left[\mathbf{Y}^{(1)}, \mathbf{Y}^{(2)},\dots,\mathbf{Y}^{(M)}\right]$, where $\mathbf{Y}^{(m)} \in \R^{N\times D_m}$. Similarly, we represent the collection of the factor loadings $\mathbf{W} \in \R^{D\times K}$ as $\left[\mathbf{W}^{(1)}, \mathbf{W}^{(2)},\dots,\mathbf{W}^{(M)}\right]$.\\
We assess the quality of each model by computing the reconstruction error measured by the RMSE,
\begin{equation*}
    RMSE(\mathbf{Y}, \hat{\mathbf{Y}}) = \sqrt{\frac{\sum_{i=1}^{N}\norm{\mathbf{y}_i - \hat{\mathbf{y}}_i}^2}{N}},
\end{equation*}
where $\hat{\mathbf{y}}_i\in\R^D$ denotes the reconstructed sample $\mathbf{y}_i$.\\
Next, we quantify the ability of each model to recover the true underlying structure of the factor loadings by computing the precision, recall and the $F_1$ score between the inferred feature sets and the true feature sets. Here, the inferred feature sets refer to the binarized representation of the factor loadings $\hat{\mathbf{W}}$ in terms of active and inactive features based on a threshold, where factor loadings with an absolute value larger than the threshold are considered active, and, otherwise inactive. 
With a slight abuse in notation, let $\mathbf{I}_{k}$ be the true feature set, and $\hat{\mathbf{I}}_{k}$ the inferred feature set for factor $k$, where $\hat{I}_{j, k} = 1$ if $\abs{\hat{w}_{j, k}} > t$, and $\hat{I}_{j, k} = 0$ otherwise, for a given threshold $t$.
Then, we may generate the confusion matrix in Table~\ref{tab:conf_matrix} to compute the precision, recall and the $F_1$ score. 
\begin{table}[t]
\caption{A confusion matrix between the inferred feature sets and the true feature sets.} \label{tab:conf_matrix}
\begin{center}
\vspace{.3in}
\begin{tabular}{@{}ccc@{}}
\toprule
 & \begin{tabular}[c]{@{}c@{}}actual active\\ $\abs{w_{j,k}} > t$\end{tabular} & \begin{tabular}[c]{@{}c@{}}actual inactive\\ $\abs{w_{j,k}} \leq t$\end{tabular} \\ \midrule
\begin{tabular}[c]{@{}c@{}}predicted active\\ $\abs{\hat{w}_{j,k}} > t$\end{tabular} & $TP$ & $FP$ \\ \midrule
\begin{tabular}[c]{@{}c@{}}predicted inactive\\ $\abs{\hat{w}_{j,k}} \leq t$\end{tabular} & $FN$ & $TN$ \\ \bottomrule
\end{tabular}
\vspace{.2in}
\end{center}
\end{table}

\begin{equation*}
    Pr = \frac{TP}{TP + FP},
    Rc = \frac{TP}{TP + FN},
    F_1 = \frac{2\cdot Pr \cdot Rc}{Pr + Rc}.
\end{equation*}

Finally, we describe the process of factor matching to overcome the issue of non-identifiability that is generally present in factor analysis. We rely on the cosine similarities between the inferred feature sets and the true feature sets to match the latent dimensions.  Specifically, we compute a cosine similarity matrix $\mathbf{C}\in\R^{K\times K}$, where $c_{k, l} = cos(\hat{\mathbf{I}}_{k},\mathbf{I}_{l})$, for $k, l \in \{1,\dots,K\}$, and compute an optimal permutation matrix $\mathbf{P}\in\R^{K\times K}$, such that the transformation $\mathbf{CP}$ has the largest values in the diagonal, i.e.\ maximizes the trace $Tr(\mathbf{CP})$. We then inspect the permutation matrix $\mathbf{P}$ to match the factor dimensions between the inferred and the actual latent space.

% \newpage
\subsection{Precision and Recall Curves}
\begin{figure*}[ht]
  \centering
  % \vspace{.1in}
  \includegraphics[width=1.0\linewidth]{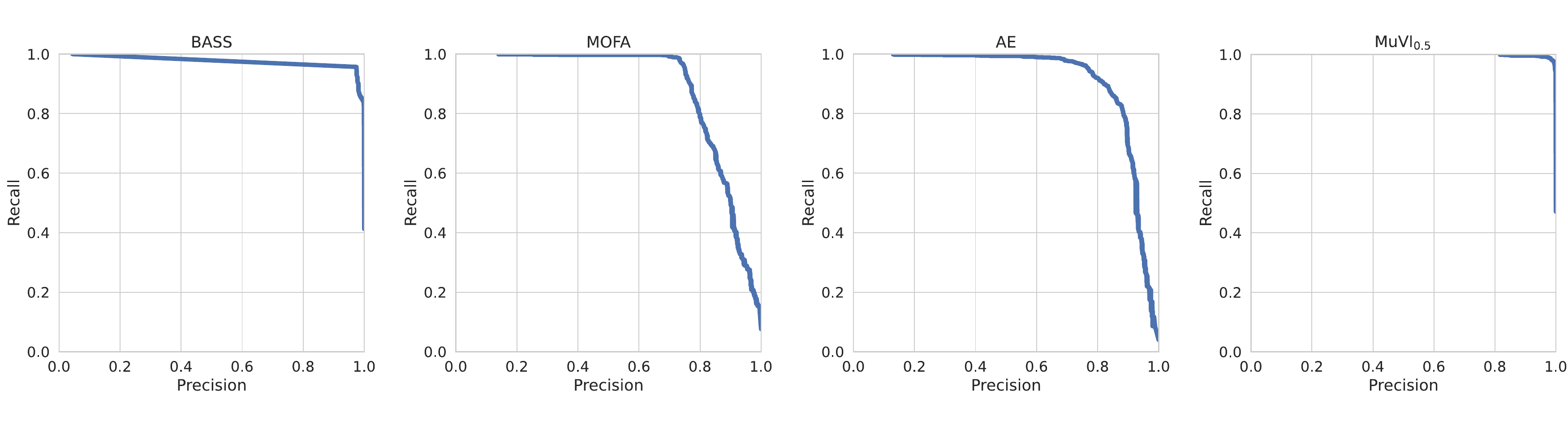}
  % \vspace{.1in}
  \caption{Threshold-agnostic precision and recall curves for MuVI$_{0.5}$ and the baseline models: BASS, MOFA, AE.}
  \label{fig:pr_rec}
\end{figure*}

% \section{APPENDIX B}\label{appendix_b}
In addition to the threshold-based $F_1$ scores provided in~\ref{subsubsec:eval}, we show the precision and recall curves across all possible thresholds in Figure~\ref{fig:pr_rec} for MuVI and the baseline models: BASS, MOFA and AE.
\newpage
\subsection{Performance in Higher Dimensions}
We repeat the synthetic experiments by exacerbating the gap between the number of samples and the number of features ($N \ll D$). For each of the four views $m$, we increase the corresponding number of features $D_m$ to one, five and ten thousand features, while keeping the number of samples low, $N=200$. The results reported in Table~\ref{tab:muvi_dim} for the uninformed version of our model (MuVI$_U$) are consistent with the results shown in~\ref{subsubsec:eval}.
\begin{table}[t]
\caption{Performance evaluation when increasing the number of features $D_m$ for each of the four views $m$, while keeping the number of samples low, $N=200$. Average scores across five independent runs $\pm$ standard deviation.} \label{tab:muvi_dim}
\begin{center}
\vspace{.1in}
\resizebox{\linewidth}{!}{%
\begin{tabular}{@{}lcccc@{}}
\toprule
\multicolumn{1}{c}{} & RMSE           & Precision      & Recall         & $F_1$       \\ \midrule\midrule
$D_m=1K$ & $0.322 \pm 0.00$ & $0.937 \pm 0.03$          & $0.995 \pm 0.00$          & $0.965 \pm 0.02$\\ \midrule
$D_m=5K$ & $0.323 \pm 0.00$ & $0.943 \pm 0.02$          & $0.995 \pm 0.00$          & $0.968 \pm 0.01$\\ \midrule
$D_m=10K$ & $0.323 \pm 0.00$ & $0.948 \pm 0.02$          & $0.996 \pm 0.00$          & $0.971 \pm 0.00$\\ \bottomrule
\end{tabular}
}%
\end{center}
\end{table}

% \newpage

\section{APPENDIX B}\label{appendix_b}
Here, we further investigate the interplay between the quality of the prior information, and the degree of the prior belief encoded by the single hyperparameter $\alpha$. 

\subsection{Sensitivity to the Prior Penalty}
\begin{figure*}
  \centering
  \vspace{.2in}
  \includegraphics[width=1.0\linewidth]{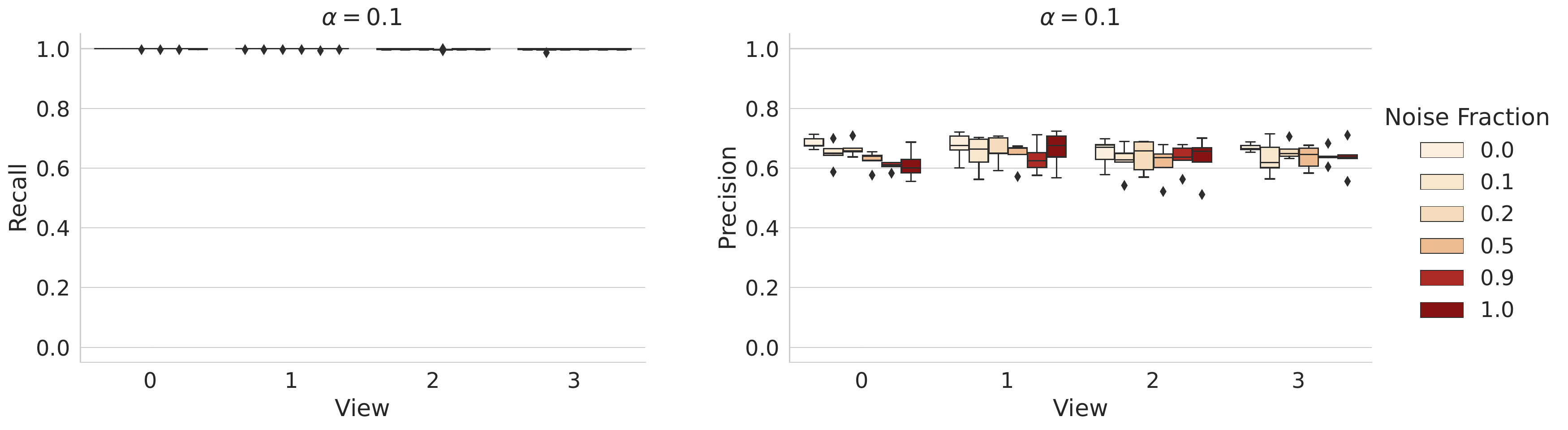}
  \includegraphics[width=1.0\linewidth]{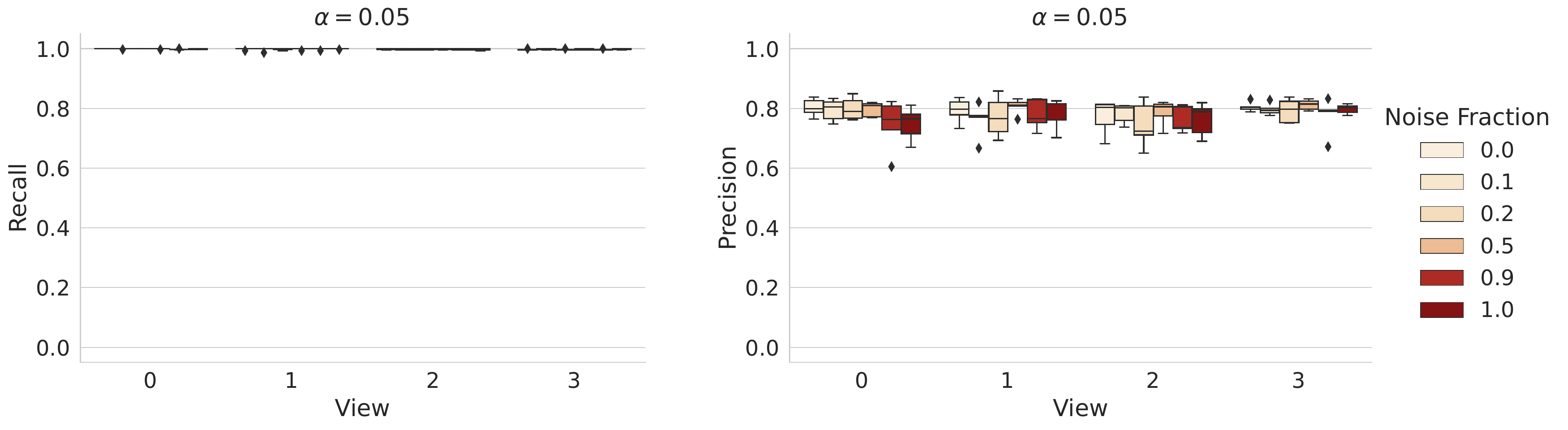}
  \includegraphics[width=1.0\linewidth]{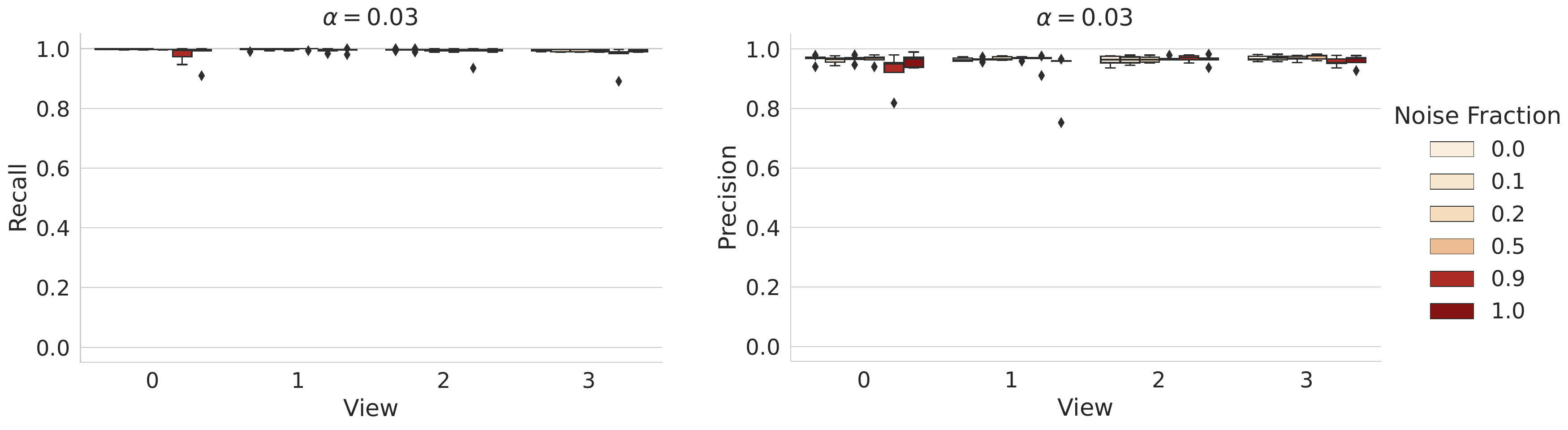}
  \includegraphics[width=1.0\linewidth]{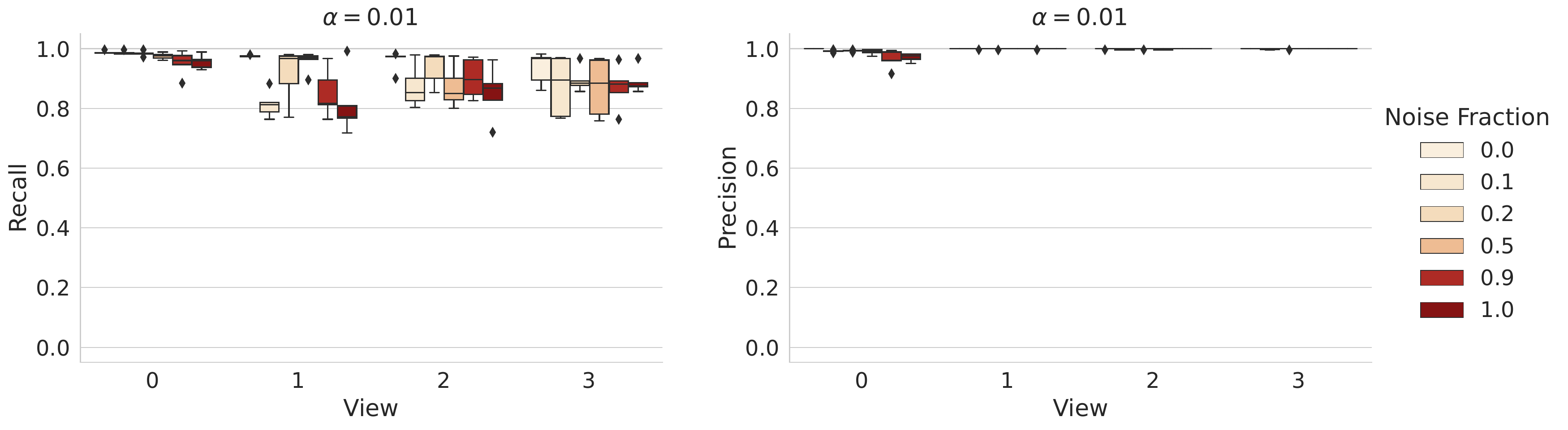}
  \vspace{.2in}
  \caption{Sensitivity analysis for the choice of $\alpha$ across different noise levels, in terms of the recall (left) and precision (right) between the inferred and the true feature sets. A larger $\alpha$ (top) translates to a weaker penalty to the features not present in the (noisy) feature set, whereas a smaller alpha (bottom) encodes a stronger prior belief, and therefore a larger penalty. Results when informing only the first view.}
  \label{fig:sens}
\end{figure*}

\begin{table}[t]
\caption{Performance comparison across all noise levels and penalties in the ablation study. Average RMSE across five independent runs $\pm$ standard deviation.} \label{tab:sens_alpha_rmse}
\begin{center}
\vspace{.1in}
\resizebox{\linewidth}{!}{%
\begin{tabular}{@{}ccccc@{}}
\toprule
 & $\alpha=0.1$ & $\alpha=0.05$ & $\alpha=0.03$ & $\alpha=0.01$ \\ \midrule
MuVI$_{0.0}$ & $0.321\pm 0.00$ & $0.321\pm 0.00$ & $0.322\pm 0.00$ & $0.328\pm 0.02$ \\ \midrule
MuVI$_{0.1}$ & $0.321\pm 0.00$ & $0.321\pm 0.00$ & $0.322\pm 0.00$ & $0.331\pm 0.03$ \\ \midrule
MuVI$_{0.2}$ & $0.321\pm 0.00$ & $0.321\pm 0.00$ & $0.322\pm 0.00$ & $0.335\pm 0.02$ \\ \midrule
MuVI$_{0.5}$ & $0.321\pm 0.00$ & $0.321\pm 0.00$ & $0.322\pm 0.00$ & $0.339\pm 0.03$ \\ \midrule
MuVI$_{0.9}$ & $0.321\pm 0.00$ & $0.322\pm 0.00$ & $0.326\pm 0.01$ & $0.367\pm 0.03$ \\ \midrule
MuVI$_{1.0}$ & $0.321\pm 0.00$ & $0.322\pm 0.00$ & $0.328\pm 0.01$ & $0.393\pm 0.03$ \\ \bottomrule
\end{tabular}
}%
\end{center}
\end{table}
Figure~\ref{fig:sens} summarizes the ablation study for the choice of $\alpha$ across different noise levels, in terms of the recall (left) and precision (right) between the inferred and the true feature sets, when informing only a single view (view 0). 
% We reiterate that we use the same threshold $t=0.1$, matching the true threshold, across all training instances for a fairer comparison. 
A larger $\alpha$ (top) translates to a weaker penalty to the features not present in the (noisy) feature set, whereas a smaller alpha (bottom) encodes a stronger prior belief, and therefore a larger penalty. As the prior belief increases, i.e.\ $\alpha$ decreases, we observe an increase in the amount of false negatives as the noise level increases. This behavior is to be expected, since the noisy prior incorrectly penalizes the majority of the true signals. In the bottom row, in particular, the high penalty induced for the uninformed views, i.e. view 1, 2 and 3, reduces their learning capacity relative to the informed view, which exhibits no difficulties in inferring the true signals. At the same time, the amount of false positives decreases, resulting in a higher precision score when having a higher degree of belief in the prior information. In other words, a weaker belief may lead to additional and redundant signals escaping the regularized horseshoe penalty, and a stronger belief may restrict some of the true signals, depending on the amount of noise in the prior information. The ablation study indicates an $\alpha\approx 0.03$ performs the best in terms of both precision and recall, and strikes a better balance between recovering the true positive signals while restricting the true negative signals.
Nevertheless, the quality of the latent representation is preserved across all noise levels and penalties. In Table~\ref{tab:sens_alpha_rmse} we report the reconstruction error for all training configurations involved in the ablation study. The RMSE obtained for the majority of the models still outperforms the baselines, and only drops slightly when combining highly perturbed prior feature sets with a high penalty.

% \newpage
\subsection{Learning Identifiable Factors from Pure Variables}
\begin{figure}[ht]
  \centering
  \vspace{.3in}
  \includegraphics[width=0.9\linewidth]{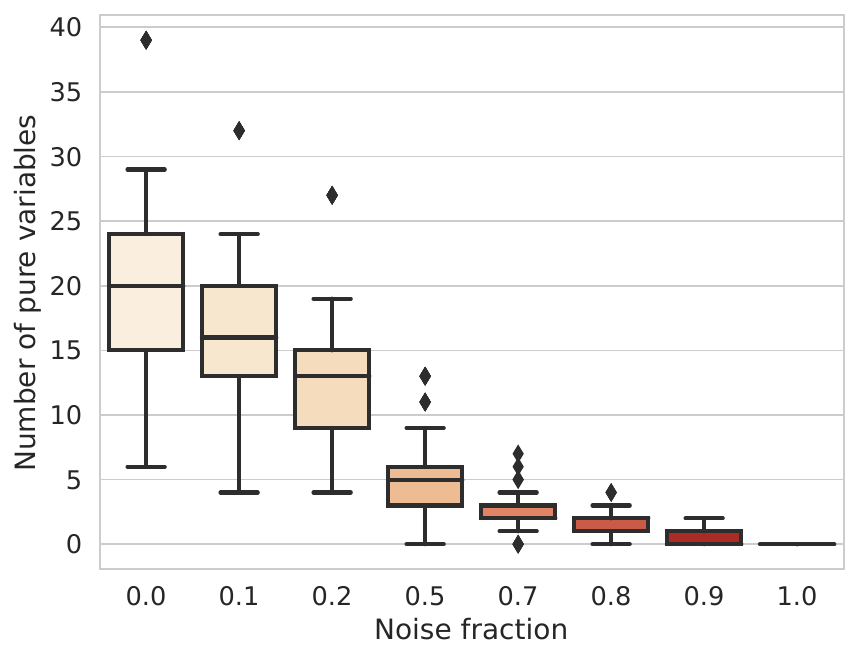}
  \vspace{.1in}
  \caption{Noise fraction as a proxy for the number of pure variables. Each boxplot summarizes the distribution of the pure variables across all factors for the feature sets generated in the synthetic experiments, for different noise levels.}
  \label{fig:pure_vars_boxplot}
\end{figure}

In Section~\ref{subsubsec:res}, in Figure~\ref{fig:boxplot_noise_new} we assess to what extent our model preserves the identifiability properties when increasing the amount of noise in the prior annotations. The noise level serves as a proxy for the number of pure variables in the prior information, that is, a higher noise level results in fewer pure variables a priori. A noise fraction of $100\%$, for instance, results in zero pure variables, and therefore makes the model unidentifiable. In Figure~\ref{fig:pure_vars_boxplot} we quantify the number of known pure variables corresponding to each noise fraction. The results are aggregated across all feature sets generated for the synthetic datasets (see~\ref{sec:exp}). 

\begin{figure}[t]
  \centering
  \vspace{-5in}
  \includegraphics[width=0.9\linewidth]{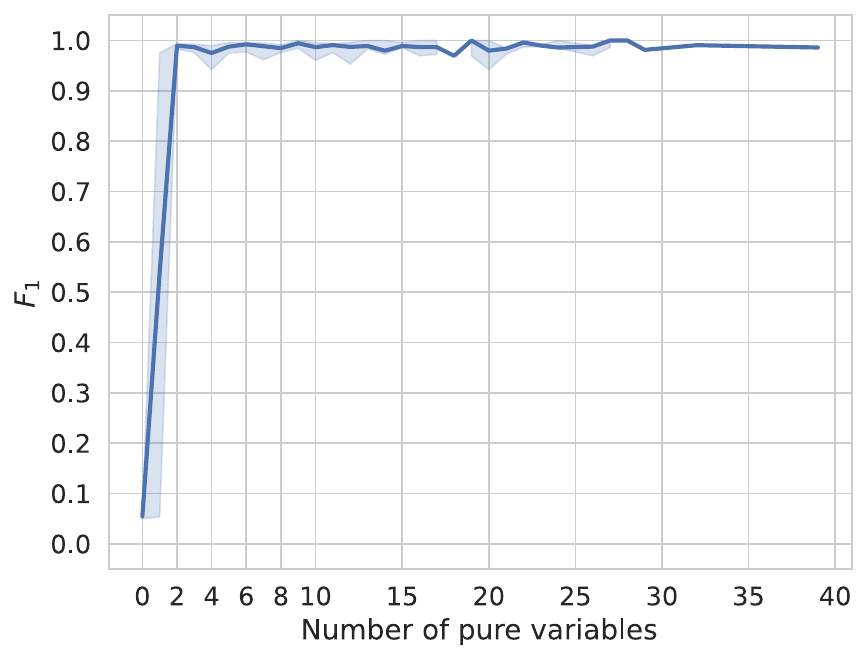}
  \vspace{.2in}
  \caption{Inferring identifiable factors from pure variables. The median $F_1$ score when increasing the number of pure variables available in the prior feature set, shaded by the $95\%$ confidence interval. The inferred and true factors are compared along the same dimension, without performing factor matching.}
  \label{fig:pure_vars_scores}
\end{figure}

Next, we investigate the minimal number of pure variables required in a feature set to render the corresponding latent factor identifiable. 
In Figure~\ref{fig:pure_vars_scores} we count the number of the pure variables and compute the $F_1$ score for each factor individually, without performing factor matching. In other words, we compare the inferred feature set $\hat{\mathbf{I}}_k$ with the true feature set $\mathbf{I}_k$ for the same latent dimension $k$. Since several factors may have the same number of pure variables, we report the median scores and a $95\%$ confidence interval. The results strongly suggest that as few as two pure variables are sufficient to guarantee identifiable factors, in accordance with the theory on the identifiability of factor analysis~\citep{anderson1956statistical}.

% \newpage
\clearpage

\section{APPENDIX C}\label{appendix_c}
We complement the analysis on the real TCGA dataset from Section~\ref{subsec:tcga} with additional results. 
\subsection{Availability of Pure Variables in the Gene Sets}
\begin{figure}[h]
  \centering
  \vspace{.2in}
  \includegraphics[width=1.0\linewidth]{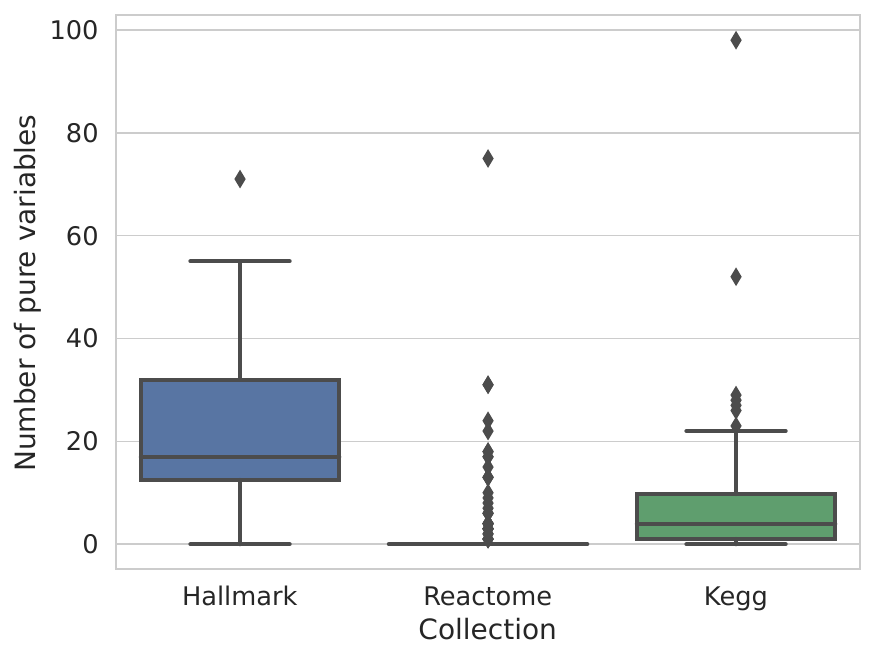}
  \vspace{.1in}
  \caption{Number of pure variables in each gene set (feature set) collection integrated in the TCGA analysis.}
  \label{fig:tcga_pv}
\end{figure}

In Figure~\ref{fig:tcga_pv}, we provide a summary of the number of pure variables available in each feature set (gene set) collection. Due to being more specialized, Hallmark and KEGG include more pure variables than the more general collection of Reactome. 
% Compared to the results in Figure~\ref{fig:pure_vars_boxplot}, Hallmark with a median number of pure variables of 17 more closely resembles the synthetic experiments with $10\%$ noise, and Kegg with $50\%$ noise. 
However, we may only pinpoint the pure variables according to the definition of the gene sets. In practice, we make no assumptions about the gene sets or the number of pure variables available in each gene set, and treat them as partially correct prior information, thereby allowing the model to refine each pathway based on the training data.
% make no assumptions whether a gene in a gene set is a pure variable or not, due to the noisy nature of these feature sets. 
\\
In order to determine whether the inferred factor loadings differ significantly from the features present in the prior gene set annotation, we perform an adjusted parametric t-test~\citep{frost2015principal} based on the correlation between the features that are present in the prior annotation.

\subsection{Latent Space Inferred by the Baselines}

\begin{figure*}[ht]
  \centering
  \vspace{.3in}
  \includegraphics[width=.33\linewidth]{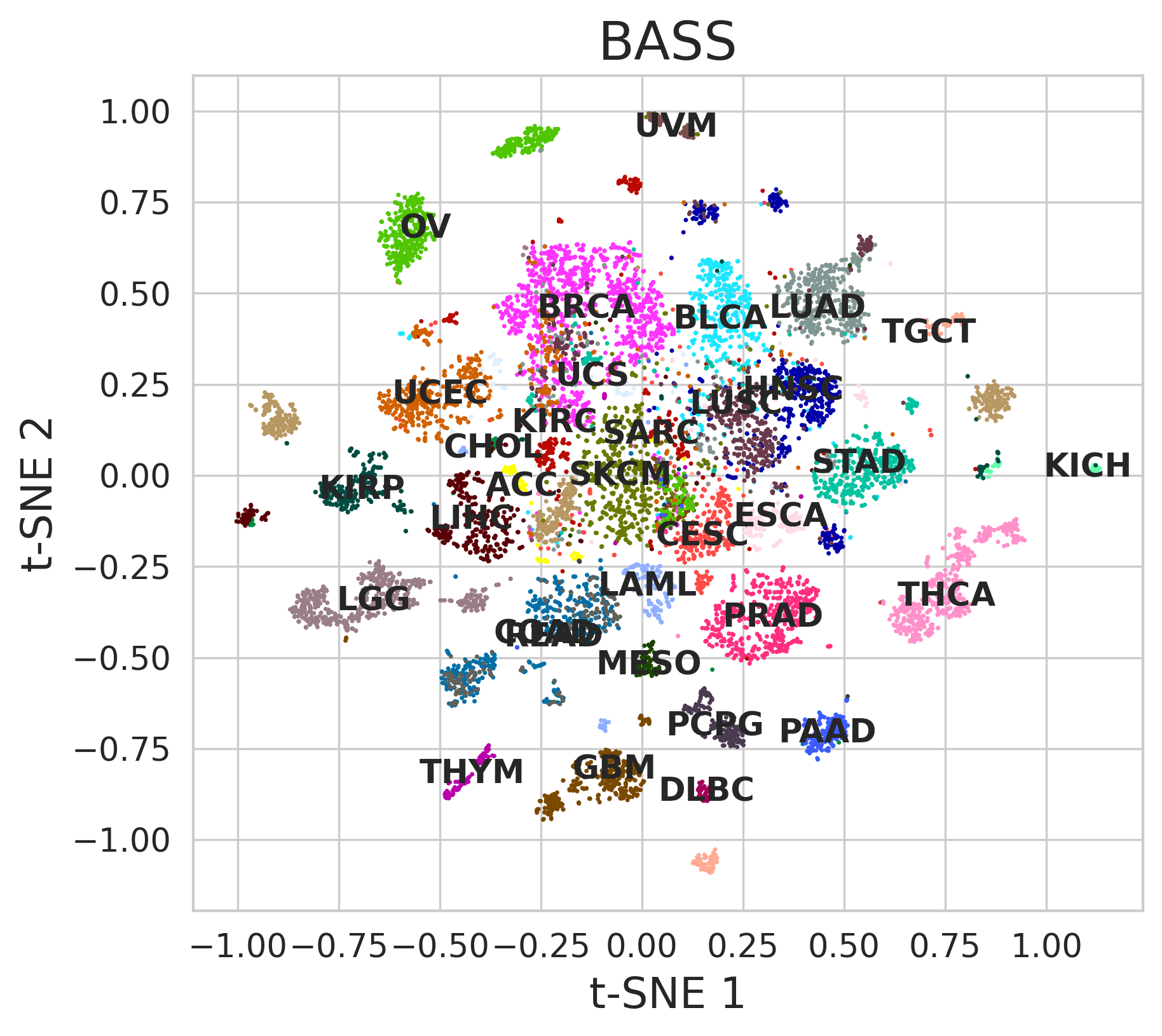}
  \includegraphics[width=.33\linewidth]{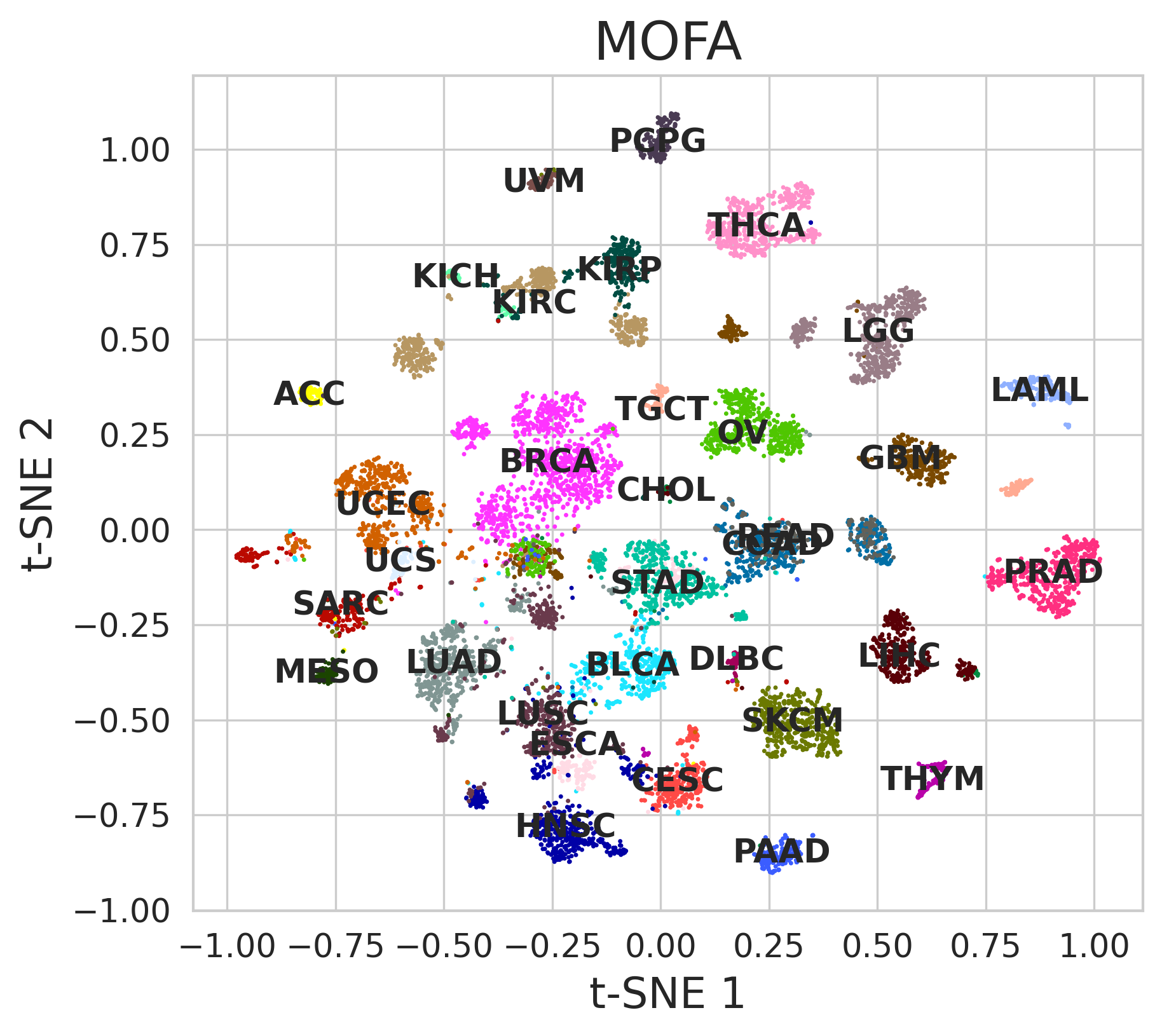}
  \includegraphics[width=.33\linewidth]{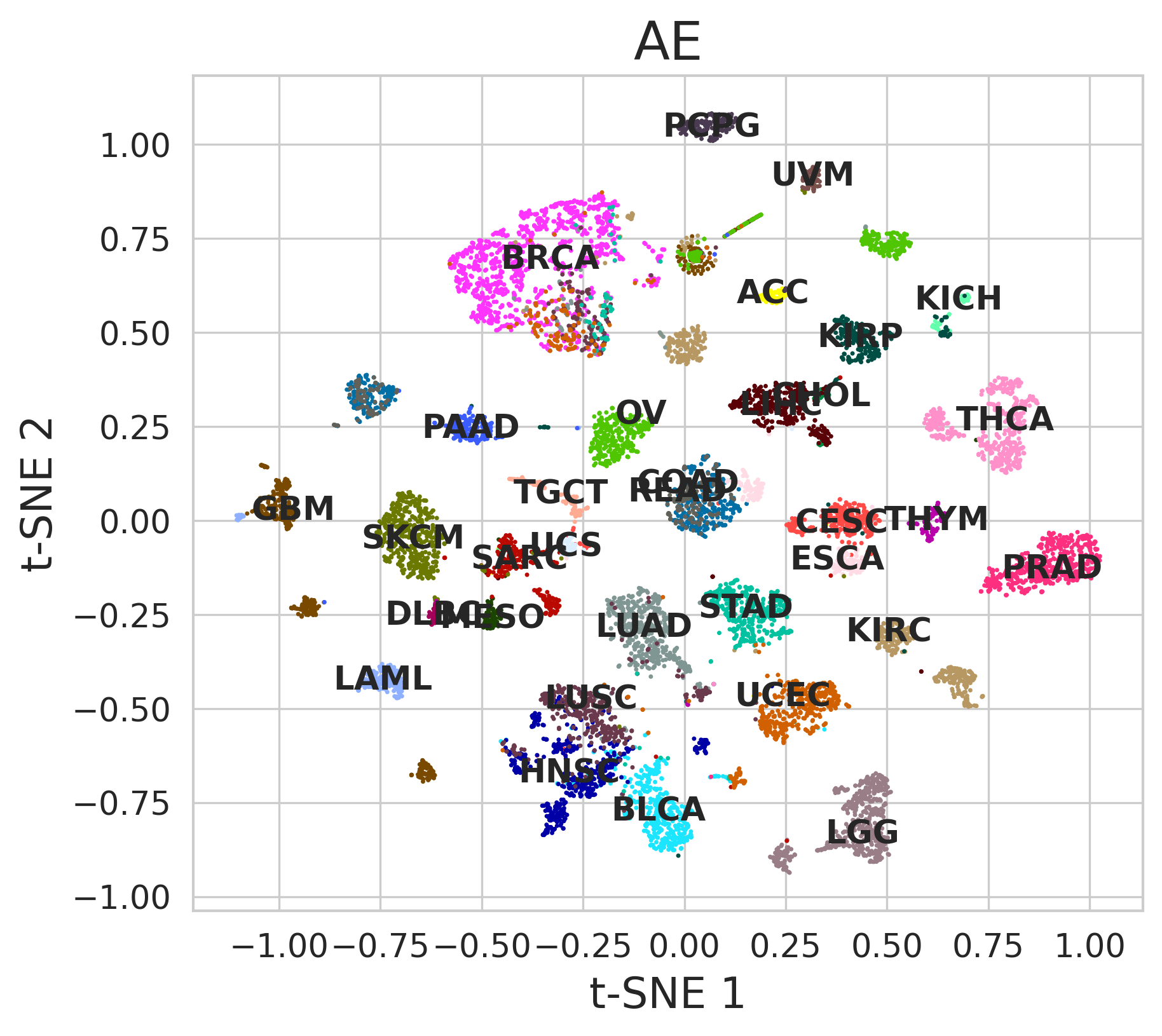}
  \vspace{.2in}
  \caption{Embedding space learned by the baselines: BASS, MOFA and the hybrid autoencoder (AE).}
  \label{fig:tcga_tsne}
\end{figure*}

For a visual comparison, we apply a t-SNE approach to the latent representations inferred by each baseline model and map the data onto the two dimensional embeddings as shown in Figure~\ref{fig:tcga_tsne}.

\subsection{Matching Relevant Factors to Cancer Types}
\begin{figure*}[t]
  \centering
  % \vspace{.1in}
  \includegraphics[width=.7\linewidth]{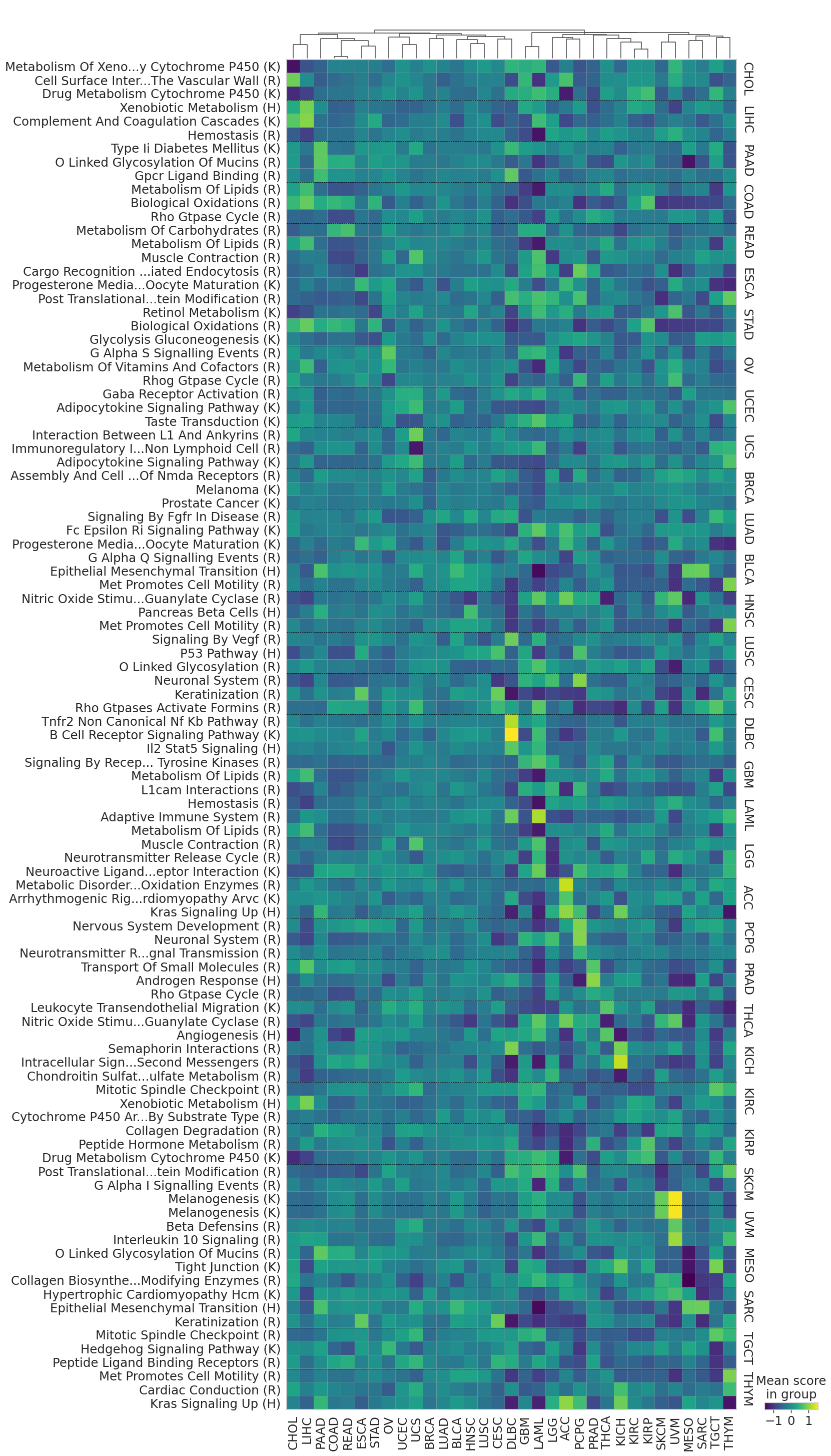}
  \vspace{.1in}
  \caption{Relevant pathways for each cancer subtype. A one-vs-rest Wilcoxon rank test identifies the main pathways that drive the heterogeneity of patients with respect to their cancer types. Each group of the top 3 pathways corresponds to a single cancer subtype on the right. The color intensity for each cell depicts the average factor score in each patient subpopulation. Cluster subtypes ordered via a standard hierarchical clustering on their factor scores.}
  \label{fig:tcga_cancer_type_test}
\end{figure*}

Finally, we provide a summary of the relevant pathways for each cancer subtype. We perform a one-vs-rest Wilcoxon rank test to identify the main pathways that capture the differences among patients of different cancer types. In Figure~\ref{fig:tcga_cancer_type_test}, we report the top 3 pathways corresponding to each cancer subtype. The test results indicate that both the androgen response and the melanogenesis pathway are highly relevant for prostate cancer (PRAD) and melanoma related cancer subtypes (SKCM and UVM). In addition, we apply a standard hierarchical clustering on the factor scores to group similar cluster subtypes together. Among others, we observe biologically meaningful structures such as identifying similarities across the only two groups of melanoma samples such as the skin cutaneous melanoma (SKCM) and uveal melanoma (UVM).
\end{document}